\documentclass{article}

\usepackage{PRIMEarxiv}

\usepackage[utf8]{inputenc} 
\usepackage[T1]{fontenc}    
\usepackage{url}            
\usepackage{booktabs}       
\usepackage{amsfonts}       
\usepackage{nicefrac}       
\usepackage{microtype}      
\usepackage{lipsum}
\usepackage{fancyhdr}       
\usepackage{graphicx}       
\graphicspath{{media/}}     

\AtBeginDocument{
    \setlength{\abovedisplayskip}{1.25em}
    \setlength{\belowdisplayskip}{1.25em}
    \setlength{\abovedisplayshortskip}{0.75em}
    \setlength{\belowdisplayshortskip}{0.75em}
}

\usepackage{amsmath}
\usepackage{amsfonts}
\usepackage{graphicx}
\usepackage{parskip}
\usepackage{comment}
\usepackage{csquotes}
\usepackage{geometry}
\usepackage{array}
\usepackage{booktabs}
\usepackage{subcaption}
\usepackage{tabularx}
\usepackage{float}
\usepackage{adjustbox}
\usepackage{listings}
\usepackage{enumitem}

\usepackage[colorlinks=true, allcolors=blue]{hyperref}

\usepackage[
backend=biber,
style=authoryear,
]{biblatex}

\def\sym#1{\ifmmode^{#1}\else\(^{#1}\)\fi}

\newcommand{\flop}{\, \textrm{FLOP}}

\newcommand{\dmodel}{d_{\text{model}}}
\newcommand{\dff}{d_{\text{ff}}}
\newcommand{\dhead}{d_{\text{head}}}
\newcommand{\dlatent}{d_{\text{latent}}}

\newcommand{\nlayer}{n_{\text{layers}}}
\newcommand{\nhead}{n_{\text{head}}}
\newcommand{\nkvhead}{n_{\text{kv head}}}
\newcommand{\nexpert}{n_{\text{expert}}}
\newcommand{\nactiveexpert}{n_{\text{active expert}}}
\newcommand{\nparam}{N_{\text{param}}}
\newcommand{\nactiveparam}{N_{\text{active param}}}

\newcommand{\npp}{N_{\text{PP}}}
\newcommand{\nep}{N_{\text{EP}}}
\newcommand{\ngpu}{N_{\text{GPU}}}

\newcommand{\microbatch}{b_{\text{micro}}}

\newcommand{\linput}{\ell_{\text{input}}}

\usepackage{enumitem}

\newcommand{\mus}{\text{ }\mu\text{s}}

\newcommand{\nreduce}{n_{\text{reduce}}}
\newcommand{\treduce}{t_{\text{reduce}}}

\title{Inference economics of language models}

\author{
  Ege Erdil \\
  Epoch \\
  \texttt{ege@epoch.ai} \\
}

\begin{document}
\maketitle

\begin{abstract} 
We develop a theoretical model that addresses the economic trade-off between cost per token versus serial token generation speed when deploying LLMs for inference at scale. Our model takes into account arithmetic, memory bandwidth, network bandwidth and latency constraints; and optimizes over different parallelism setups and batch sizes to find the ones that optimize serial inference speed at a given cost per token. We use the model to compute Pareto frontiers of serial speed versus cost per token for popular language models.
\end{abstract}

\section{Introduction}

Large language models (LLMs) have significantly influenced the field of artificial intelligence, becoming increasingly prevalent in both research and practical applications. As these models have grown and become used in a wide range of applications, their deployment for real-world tasks has precipitated a notable shift in the economics of AI. The cumulative inference costs—the total computational expenses associated with running these models across all instances—have escalated to levels comparable to, or in some cases exceeding, the initial training costs. Concurrently, a competitive market has emerged, with numerous API providers striving to offer open-source LLMs at optimal price points and processing speeds. Despite these developments, there remains a critical gap in our understanding: a comprehensive, quantitative analysis of the inference costs for LLMs when deployed at scale has yet to be conducted.

This is more than a theoretical problem. In practice, it's crucial for engineers working on serving LLMs for inference cheaply and quickly to have a sense of the limits of what's achievable with their hardware under ideal conditions. For kernel programmers who implement single-GPU kernels in CUDA, the roofline model of taking a minimum over arithmetic and memory read-write times serves this purpose: if a kernel is failing to achieve good performance relative to this baseline, we can often assume its implementation is inefficient and a theoretical best kernel should be able to do better. 

Past work on how LLMs can be served efficiently for inference (\cite{pope2022efficiently}, \cite{steinhardt2022}) is either "too practical", focusing strictly on kernel performance without comparing it to some theoretical best result, or "too theoretical", abstracting too far from concrete hardware details and therefore not providing useful guidance for real-world inference economics. In short, there's no equivalent to the "roofline model" in the literature about LLM inference right now, though we suspect AI labs and inference providers have their own internal models which they haven't released to the public.

Here are two questions, one very concrete and one more theoretical, that are difficult to answer without such a model:

\begin{itemize}
    \item Suppose an open-source API provider succeeds in serving 16-bit Llama 3 70B on a DGX H100 machine at short contexts with a serial speed of 70 tokens/second and a hardware utilization rate of 15\%. Is this performance good or bad, and is it reasonable to hope we might do better, i.e. increase either serial speed or hardware utilization without decreasing the other? 

    \item Past theoretical work on LLM inference, e.g. \cite{steinhardt2022}, has observed that because individual matrices are two-dimensional while matrix multiplications are three-dimensional, parallelizing a matrix multiplication by a factor of \( 2 \) along all its dimensions only increases required memory and network communication for parallelism by a factor of \( 2 \). Therefore, even when memory and network reads become binding, we should expect to be able to quadruple the serial speed of matrix multiplications (and thus of LLM inference) for each halving of hardware utilization by simply parallelizing inference across eight times the number of GPUs. If so, why are LLMs typically not served at much faster speeds?
\end{itemize}

Our goal is to fill this gap by building a theoretical model that can take an LLM architecture and a GPU as inputs and determine how costly it is to serve this model at a particular context length and a particular rate of tokens per second per request if an optimal inference setup is used. We will then evaluate our model by comparing its predictions against the empirical performance figures for various API providers from using data on AI language models and API providers.

The paper is structured as follows: in Section \ref{sec:motivation} we first motivate the ingredients that will go into our complete model by analyzing some of the factors determining inference economics in isolation. This section is useful to build intuition for how inference economics works and to motivate why we choose to include some components in the model while excluding others. In Section \ref{sec:model_description} we give a complete description of our model with all of its parts before using the model to make predictions about well-known models such as DeepSeek-V3 (\cite{deepseekai2025deepseekv3technicalreport}). Finally, we have a takeaways section in which we list some of the important facts about inference economics revealed by our analysis.

We will assume familiarity with basic features of the decoder-only Transformer architecture and of mixture-of-experts models, but not with the technical details of distributed inference methods.

The code needed to reproduce the results in this paper can be found in \href{https://github.com/ege-erdil/inference-economics}{this GitHub repository}.

\section{Motivation for the model}
\label{sec:motivation}

\subsection{Simplest case: single device instances, short context inference, dense model}
\label{sec:motivation-simplest-case}

The simplest case for thinking about inference economics is the case of a dense model where we're doing short context length inference and each request is routed only to a single device. This removes most of the complications that have to be included in a general model: there's no possibility of using complex distributed inference methods other than just routing requests to different instances, and the attention computations, along with KV cache reads, are assumed to be negligible because of the short context assumption. 

In this case, inference economics of a model are determined by its total number of parameters \( \nparam \), the precision of the model parameters \( p \), the FLOP per second of the GPU \( C \) and the HBM bandwidth of the GPU \( B \). If we assume a batch size equal to \( b \) requests, then we will obtain the following:
\begin{align}
    \text{token latency} &= \max \left( \frac{p \cdot \nparam}{B}, \, \frac{(2 \flop) \cdot \nparam \cdot b}{C} \right) \\
    \text{GPU seconds per token} = \frac{\text{token latency}}{b} &= \max \left( \frac{p \cdot \nparam}{b \cdot B}, \, \frac{(2 \flop) \cdot \nparam}{C} \right)
\end{align}

where in both equations, the first term in the maximum measures the time taken for parameter reads while the second term measures the time taken for arithmetic. We take a maximum assuming that memory reads and arithmetic are overlapped whenever possible.

In this simple case, the batch size \( b \) is the only variable parameter affecting the inference economics. The optimal batch size, denoted as \( b^{*} \), is given by the equation \( b^{*} = p \cdot C/(B \cdot 2 \flop) \). At this batch size, the time taken for parameter reads and the time taken for arithmetic computations are equal. Consequently, \( b^{*} \) represents the maximum batch size that allows for the minimum token latency of \( p \cdot \nparam/B \) to be achieved.

When not limited by an insufficient number of incoming requests, it is always optimal to set the batch size \( b \) equal to \( b^{*} \). If the number of incoming requests exceeds \( b^{*} \), it is more efficient to distribute the requests across multiple instances rather than increasing the batch size on a single instance. This is because increasing the batch size beyond \( b^{*} \) results in higher token latency without improving the cost in terms of GPU seconds per token. Conversely, reducing the batch size below \( b^{*} \) leads to increased cost without any improvement in token latency.

In situations where the number of incoming requests is insufficient to achieve a batch size of \( b^{*} \) while maintaining a token latency of \( p \cdot \nparam/B \) on a single device, the optimal approach is to set the batch size \( b \) as close to \( b^{*} \) as possible given the available requests.

\subsection{Multiple device instances with infinite network bandwidth, short context inference, dense model}
\label{sec:model-stage-2}

In this section, we introduce an additional free parameter, the instance size (\( \ngpu \)), which represents the number of GPUs utilized in the system. To simplify the analysis, we assume a network with nonzero latency but infinite bandwidth, allowing us to focus on the effects of distributing the workload across multiple GPUs without considering the impact of the number of words being reduced across different GPUs.

We'll assume that we use tensor parallelism across all of the GPUs in one instance: data parallelism is already taken into account because of the batch size being a free parameter, and in most cases pipeline parallelism is going to be inferior to data parallelism and so we can assume that it is not being used. (The only case where this is not true is when memory constraints are binding for otherwise efficient inference setups, which is rarely the case on NVIDIA hardware but is much more relevant when using specialized hardware such as Groq LPUs with very small amounts of memory per chip, as we'll see later.)

Utilizing multiple GPUs enables the distribution of arithmetic operations and parameter reads, potentially reducing the time spent on these tasks. However, this comes at the cost of introducing network latency due to the necessity of performing all-reduce operations in each layer. The token latency now depends on several factors:
\begin{itemize}
    \item The number of model layers (\( \nlayer \))
    \item The number of sequential all-reduce operations required per layer (\( \nreduce \))
    \item The average latency of performing an all-reduce operation (\( \treduce \)).
\end{itemize}

Considering these variables, a simple model for token latency and GPU seconds per token can be expressed as follows:
\begin{align}
    \text{token latency} &= \nlayer \cdot \nreduce \cdot \treduce + \max \left( \frac{p \cdot \nparam}{\ngpu \cdot B}, \, \frac{(2 \flop) \cdot \nparam \cdot b}{\ngpu \cdot C} \right) \\
    \text{GPU seconds per token} &= \frac{\ngpu \cdot \nlayer}{b} \cdot \nreduce \cdot \treduce + \max \left( \frac{p \cdot \nparam}{b \cdot B}, \, \frac{(2 \flop) \cdot \nparam}{C} \right)
\end{align}

Nevertheless, this simple model has several shortcomings:

\begin{enumerate}
    \item We assume that memory reads cannot be overlapped with network latency: devices are just waiting to receive information from the previous stage in order to start reading information from memory for the current stage. However, it's possible that devices can load some fraction of the weights from HBM into the L2 cache in advance. This fraction is usually negligible for a single device because the L2 cache is usually quite small, but with multiple devices this strategy can actually be quite effective.\footnote{For example, Llama 2 70B (\cite{touvron2023llama}) has 70 billion parameters and 80 layers at 16-bit precision, so each layer is around 1.75 GB in size. With 50 MB of L2 cache per GPU, we only need 35 GPUs to fit the parameters in each layer into L2. An instance with more GPUs than this could therefore overlap most of the network latency with parameter reads from HBM.}

    \item We assume that \( \treduce \) is constant, but in general, \( \treduce \) is an increasing function of the instance size and also depends on which network topology is used to link the GPUs together. For instance, GPUs linked in a ring topology will have poor all-reduce latency at large instance sizes compared to those linked in a tree topology because each all-reduce will need to travel around the ring twice, once for the reduce-scatter and another time for the all-gather operation. The scaling of this nonzero network latency with the instance size \( \ngpu \) turns out to be the key factor that stops us from indefinitely quadrupling speed for each halving of hardware utilization in the \( \ngpu \to \infty \) limit.
\end{enumerate}

We will ignore point (1) in the rest of this work for simplicity and focus only on addressing point (2), which is far more quantitatively significant.

\subsubsection{Intra-node and network latency}

The latency of an all-reduce operation depends on two factors: the communication latency between GPUs and the number of communications required. For instance, if we have a ring topology of size \( R \) with a maximum communication latency of \( t_{\text{comm}} \) across two adjacent GPUs in the ring, the theoretical ring all-reduce latency will be \( 2(R-1) t_{\text{comm}} \) because we need a total of \( R-1 \) hops in the reduce-scatter phase and another \( R-1 \) hops in the all-gather phase. 

Unfortunately, \( t_{\text{comm}} \) is in practice not a simple constant but depends on many details: the network topology, the hardware used for communication and the exact communication protocol, as well as whether communication can be confined to a single node or must occur across node boundaries. For now, we will neglect these details and assume that \( t_{\text{comm}} \approx 1 \mus \) on the basis of information from \cite{jeaugey2019massively}, assuming that we use the lowest latency communication protocol available in NCCL (which is appropriately called ``LL"). This is about the best that can be done unless we resort to using the more detailed latency calculations from the NCCL codebase.\footnote{NVIDIA provides their own model for estimating the latency of GPU primitives at the official GitHub repository for NCCL (\cite{NCCL}) in the file ``src/graph/tuning.cc". The variable containing the hardware latency information that we need is called ``hwLat" and has latency figures in units of microseconds.} We leave this work for the next section.

We now turn to characterizing the number of participants \( R \): in general, when \( \ngpu \) is large the number of participants in each all-reduce operation in a forward pass scales as \( R \propto \ngpu^{1/2} \). This is because it's optimal for both memory bandwidth and network bandwidth optimization to slice weight matrices close to evenly along both dimensions in the large \( \ngpu \) regime. Because each all-reduce happens across the intermediate dimension \( K \) of a matrix multiplication of type \( M \times K \times N \), the number of participating ranks is the number of ways the \( K \) dimension is sliced, which will scale proportionally to \( \ngpu^{1/2} \).

Combining all of this information, we might approximate \( t_{\text{reduce}} \approx 2 \mus \cdot (\ngpu^{1/2} - 1) \) for values of \( \ngpu \) that are not too large (because with a sufficiently large instance size we would switch to using e.g. a tree topology, which would have lower latency) and at small enough batch sizes that network bandwidth constraints are not binding even when the LL algorithm is used. This expression is enough to imply an interesting fact: there is an \textit{optimal instance size} to be serving any model at if we want to minimize token latency \textit{even if we do not care about cost}. This is readily apparent from the expression for token latency that we've obtained:

\begin{equation}
    \text{token latency} = \nlayer \cdot \nreduce \cdot 2 \mus \cdot (\ngpu^{1/2} - 1) + \max \left( \frac{p \cdot \nparam}{\ngpu \cdot B}, \, \frac{(2 \flop) \cdot \nparam \cdot b}{\ngpu \cdot C} \right)
\end{equation}

The only parameter we have not characterized so far is the number of sequential all-reduces per layer \( \nreduce \). When \( \ngpu \) is large, this will be equal to the number of serial matrix multiplications in each layer\footnote{There are special cases in which this number is less: for example, when \( \ngpu \) is small we can use 1D tensor parallelism in the feedforward block and avoid the all-reduce over the \( \dff \) dimension, and we might also be able to afford partitioning attention over heads instead of over batch and avoid unnecessary all-to-all communication in that step as well. \( \nreduce = 4 \) assumes we're not in the regime where these optimizations would apply.} In a traditional Transformer operating at short context lengths, the attention mechanism itself is negligible, so we're left with \( \nreduce = 4 \) sequential matrix multiplications:

\begin{enumerate}
    \item The computation of the query, key and value vectors.
    \item The post-attention projection.
    \item The \( \dmodel \to \dff \) matrix multiplications in the feedforward block. Note that there could be more than one of these in an activation scheme such as SwiGLU (\cite{shazeer2020glu}), but they can in principle be parallelized and so don't count as multiple serial operations.
    \item The \( \dff \to \dmodel \) matrix multiplications in the feedforward block.
\end{enumerate}

While this particular arrangement of a single layer remains common today, it's not the only available option. Some architectures such as the PaLM models from \cite{chowdhery2022palm} may opt to use \textit{parallel attention}, which changes the formulation of a Transformer layer (ignoring the layer normalization operations etc.) from Equation \ref{eq:sequential-attention} to Equation \ref{eq:parallel-attention}.
\begin{align}
    \label{eq:sequential-attention}
    \text{Sequential attention: } x_{\text{output}} &= x_{\text{input}} + \text{Feedforward}(\text{Attention}(x_{\text{input}})) \\
    \label{eq:parallel-attention}
    \text{Parallel attention: } x_{\text{output}} &= x_{\text{input}} + \text{Feedforward}(x_{\text{input}}) + \text{Attention}(x_{\text{input}})
\end{align}
The advantage of this change is that it removes the logical dependence of the feedforward layer outputs on the attention layer outputs, allowing the two layers to be executed in parallel. This can potentially reduce \( \nreduce \) from \( 4 \) to \( 2 \), which is a minor gain during training but a significant gain during inference due to the tighter latency constraints there.

\subsubsection{Predictions}
\label{sec:simplified-predictions}

This model is already rich enough to make useful predictions about the token economics of dense models. To do this, we compute the token latency and the GPU seconds per token across a grid of values for the batch size \( b \) and the instance size \( \ngpu \), impose an overall throughput constraint
\[ b/(\text{token latency}) \leq \text{token throughput} \]
to account for the possibility of limited demand, convert the cost from units of GPU-seconds to units of dollars by multiplying by an hourly rental rate per GPU, and plot the implied Pareto frontiers for the per-request token generation rate on the x-axis versus the cost per million tokens on the y-axis.

\begin{figure}[h]
\centering
  \includegraphics[width=0.6\linewidth]{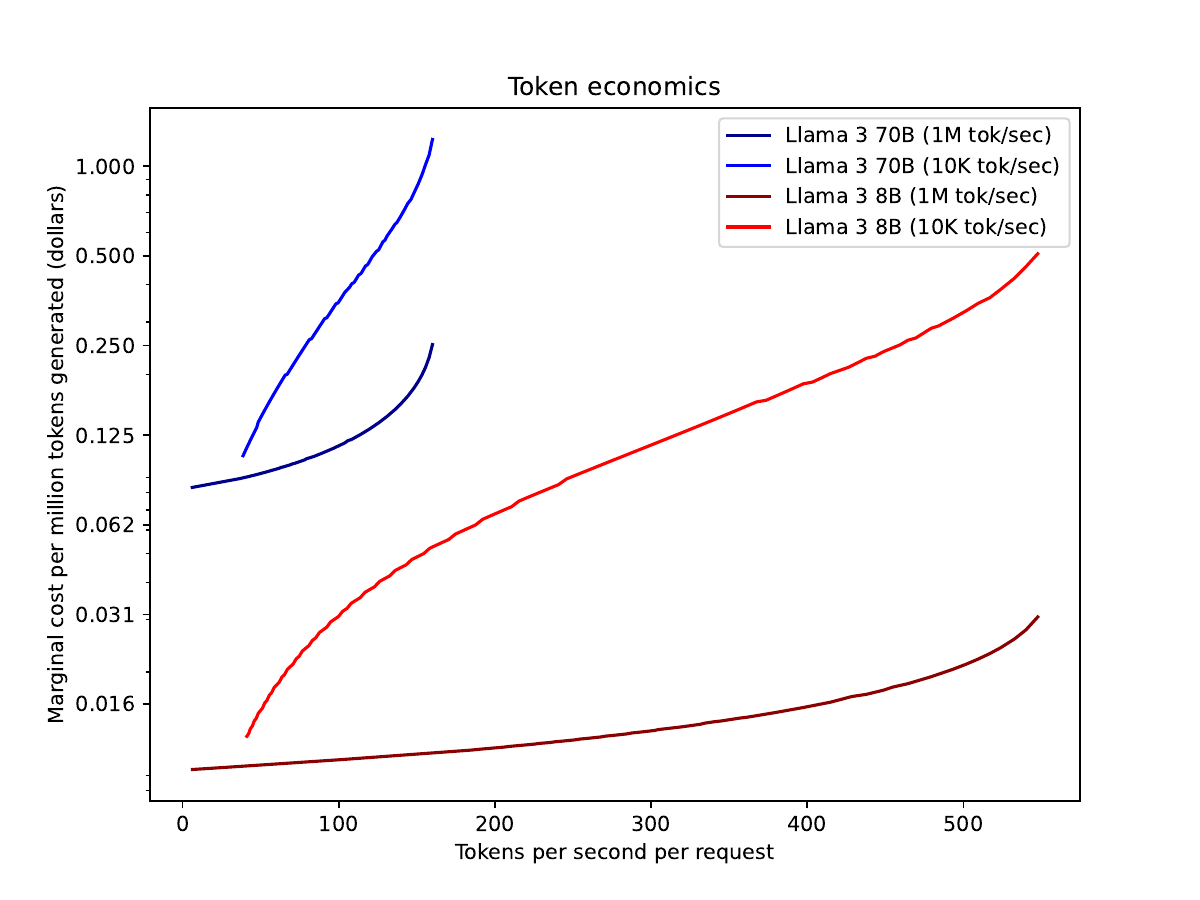}
  \caption{The token economics for Llama 3 8B and Llama 3 70B implied by our current model. We assume the H100 GPU is used at an hourly rental rate of \( \$2/\text{hour} \) per device. In the legend, "tok/sec" refers to the \textit{total demand} for the model across all users and not the token generation rate seen by a single user. In contrast, the x-axis refers to the rate seen by a single user after they submit a request and not the overall throughput.}
  \label{fig:llama-simple-token-latency-plot}
\end{figure}

The model in this section is crude, neglecting many important details, but as Figure \ref{fig:llama-simple-token-latency-plot} shows it can already roughly match empirical data. If we exclude Groq (since they use their own custom SRAM-based hardware for inference instead of NVIDIA GPUs), \cite{ArtificialAnalysis2024} reports a maximum inference speed of \( \approx 400 \text{ tokens/second} \) at 20 cents per million tokens for Llama 3 8B and \( \approx 150 \text{ tokens/second} \) at 90 cents per million tokens for Llama 3 70B, both achieved by Fireworks. In both cases, the models appear to be served within a factor of \( 2 \) of the maximum tokens per second and markups charged by API providers over just the GPU rental costs appear to be less than a factor of \( 2 \). 

\subsubsection{Theoretical analysis}
\label{sec:theoretical-analysis}

Though more complex than the model from Section \ref{sec:motivation-simplest-case}, this model is still simple enough for us to obtain analytic results about it. We repeat the token latency equation with the appropriate substitutions here for completeness:
\begin{equation}
    \text{token latency} = \nlayer \cdot \nreduce \cdot t_{\text{hop}} \cdot 2(\ngpu^{1/2} - 1) + \max \left( \frac{p \cdot \nparam}{\ngpu \cdot B}, \, \frac{(2 \flop) \cdot \nparam \cdot b}{\ngpu \cdot C} \right)
\end{equation}
The goal of an API provider is to choose \( \ngpu, b \) to minimize token latency for a given per-token cost $c$, which can be expressed as \( \text{token latency} \cdot \ngpu/b = c \).

We first eliminate one of the variables, namely \( b \). As in Section \ref{sec:motivation-simplest-case}, the efficient unconstrained choice for \( b \) is always \( b^{*} = p \cdot C/(B \cdot 2 \flop) \). This makes the memory bandwidth and arithmetic terms inside the maximum equal to one another and it's easy to see it achieves both the smallest possible cost and the smallest possible token latency for any fixed value of \( \ngpu \). If we assume \( b = b^{*} \), we can simplify the expression for token latency to
\begin{equation}
    \label{eq:simplified-token-latency}
    \text{token latency} = \nlayer \cdot \nreduce \cdot t_{\text{hop}} \cdot 2(\ngpu^{1/2} - 1) + \frac{p \cdot \nparam}{\ngpu \cdot B}
\end{equation}
We now turn our attention to a question which turns out to have a clean analytical answer: \textit{minimum possible token latency}. Because Equation \ref{eq:simplified-token-latency} diverges as \( \ngpu \to \infty \) and is a continuous function of \( \ngpu \), we know it must have a global minimum. To find this minimum, we differentiate with respect to \( \ngpu \) and set the result to zero to get
\begin{align}
    \nlayer \cdot \nreduce \cdot t_{\text{hop}} \cdot \ngpu^{-1/2} &= \frac{p \cdot \nparam}{\ngpu^2 \cdot B} \\
    \ngpu^{3/2} &= \frac{p \cdot \nparam}{\nlayer \cdot \nreduce \cdot t_{\text{hop}} \cdot B} \\
    \ngpu &= \left( \frac{p \cdot \nparam}{\nlayer \cdot \nreduce \cdot t_{\text{hop}} \cdot B} \right)^{2/3}
\end{align}
One caveat is that we might have that this optimal value of \( \ngpu \) is less than \( 1 \), which is going to lead to an unrealistic negative value for the latency term. In this case, the optimal solution is always \( \ngpu = 1 \), as using only a fraction of a single GPU would only increase memory read time without having any network latency benefit. Strictly speaking, therefore, the optimal solution is
\begin{equation}
    \label{eq:optimal-ngpu-choice}
    \ngpu^{*} = \max \left( \frac{p \cdot \nparam}{\nlayer \cdot \nreduce \cdot t_{\text{hop}} \cdot B}, 1 \right)^{2/3}
\end{equation}
Plugging this into the token latency expression now gives
\begin{equation}
    \label{eq:minimum-token-latency}
    \text{minimum token latency} = \begin{cases}
    3 \cdot (\nlayer \nreduce t_{\text{hop}})^{2/3} \cdot (p \nparam/B)^{1/3} - 2 \nlayer \nreduce t_{\text{hop}} & \text{if $\ngpu^{*} > 1$} \\
    p \cdot \nparam/B & \text{otherwise}
      \end{cases}
\end{equation}
The \( \ngpu^{*} > 1 \) condition is comparing two different timescales: the amount of time it takes to read the parameters of the model from HBM \textit{on a single GPU} and the amount of time spent for each all-gather or reduce-scatter operation per participant in a single forward pass. If the memory read time is greater, then it's optimal to use an instance of multiple GPUs; otherwise it's not.

We can derive another interesting result in the 
\( \ngpu^{*} \gg 1 \) regime about the Pareto frontiers of cost versus speed. The cost (in units of GPU seconds) of achieving the minimum possible latency is given by

\begin{align}
    \label{eq:cost-per-token-at-minimum-token-latency}
    \text{cost per token at minimum token latency} &= \frac{\ngpu^{*}}{b^{*}} \cdot \text{minimum token latency} \\
    &\approx 3 \cdot \frac{p \cdot \nparam}{B \cdot b^*} \text{(approx. valid because } \ngpu^* \gg 1 \text{ by assumption}) \\
    &= 3 \cdot \frac{(2 \flop) \cdot \nparam}{C}
\end{align}

In other words, achieving the \textit{best possible token latency} only costs around three times as much as the minimum possible cost per token arising from arithmetic alone. This result is not realistic: in actual inference setups the Pareto frontier of speed versus cost can range across much wider values for cost per token. This toy model, however, is useful because it demonstrates this is entirely because of network bandwidth constraints. The need to economize on it provides an incentive to reduce the batch size beyond \( b^* \) to reduce the sizes of activation tensors, which potentially reduces arithmetic intensity and trades off against the efficient use of memory bandwidth. We will see how this works concretely in later sections when we develop a more complete model taking network bandwidth constraints into account.

Computing the maximum tokens per second for some well-known models running on H100 SXM GPUs using Equation \ref{eq:minimum-token-latency} with \( t_{\text{hop}} = 1 \mus \) and \( \nreduce = 4 \) gives the results presented in Table \ref{tab:max-tokens-per-second-approx-by-model}.

\begin{table}[h]
\centering
\begin{tabular}{lcc}
\toprule
\textbf{Model name} & \textbf{Maximum tokens per second} & \textbf{Optimal instance size (GPUs)} \\ \midrule
Llama 3 8B          & 966                         & 11 \\ 
Llama 3 70B         & 234                         & 26 \\ 
GPT-3               & 148                         & 42 \\ 
PaLM-540B           & 86                          & 79 \\
GPT-4 (architecture from \cite{semianalysis_gpt4})              & 56                          & 173 \\
\bottomrule
\end{tabular}
\caption{Maximum tokens per second and the optimal instance size required to achieve the maximum according to Equation \ref{eq:minimum-token-latency} with \( t_{\text{hop}} = 1 \mus, \, \nreduce = 4 \) and using H100 SXM GPUs.}
\label{tab:max-tokens-per-second-approx-by-model}
\end{table}

It's worth emphasizing that we expect the tokens per second achieved in practice to be below the  figures reported in Table \ref{tab:max-tokens-per-second-approx-by-model} for many reasons:

\begin{enumerate}
    \item Even in theory, these are \textit{maximum achievable} figures for tokens per second. In practice, an API provider might opt to provide a model at a higher token latency to be able to serve it more cheaply, especially if the extra serial speed is not useful for the typical use cases of the model being served. This is especially true if the provider faces an overall throughput constraint, as in Figure \ref{fig:llama-simple-token-latency-plot}.

    \item Achieving results that are this good requires being able to eliminate the base latency in GPU collectives such as all-reduce and all-to-all. If this is not feasible, the maximum tokens per second will be less, especially for small models such as Llama 3 8B.

    \item Our analysis is based strictly on short-context inference in which the attention mechanism is assumed to be negligible, both for arithmetic and memory reads. We'll see later on how the figures for long-context inference can be significantly worse, especially if multi-query attention (\cite{shazeer2019fast}) is not used.

    \item We assume that \( \ngpu \) takes values in the real numbers, while in practice it not only takes integer values but must satisfy certain divisibility relations to ensure e.g. tensor parallelism can be effectively used during inference. 

    \item The approximation that \( R \approx \ngpu^{1/2} \) is mildly optimistic: in practice some all-reduces will involve more than \( \ngpu^{1/2} \) ranks while some will involve less, and due to the convexity involved this can slightly raise the token latency we expect.
\end{enumerate}

Table \ref{tab:max-tokens-per-second-approx-by-model} also shows that even for an 8 billion parameter model, we have \( \ngpu^{*} \gg 1 \), and the optimal instance size increases as the size of the models grow. This is because \( \nparam = O(\dmodel^2 \cdot \nlayer) \) typically grows much faster than \( \nlayer \), for large models we have \( p \nparam/B \gg \nlayer \nreduce t_{\text{hop}} \) and the optimal instance size for inference will be \( \ngpu \gg 1 \) (this is disregarding the HBM size of a single device, which in practice is also a constraint forcing us to use a bigger instance size when we're working with large models).

In this typical regime when \(\nparam\) is large, from equation \ref{eq:minimum-token-latency} we know that a good approximation to the minimum token latency is simply
\begin{equation}
    \label{eq:minimum-token-latency-approx}
    \text{minimum token latency} \approx 3 \cdot (\nlayer \nreduce t_{\text{hop}})^{2/3} \cdot (p \cdot \nparam/B)^{1/3}
\end{equation}

Equation \ref{eq:minimum-token-latency-approx} is quite useful for understanding the factors that slow inference down and their relative importance. The number of layers of a model matters more than its width for how quickly one can inference the model: as \( \nparam = O(\dmodel^2 \cdot \nlayer) \), the minimum token latency scales as \( \propto \nlayer \cdot \dmodel^{2/3} \) in the model depth and width, holding hardware features constant. In addition, increasing memory bandwidth also matters less than reducing the latency of GPU collective primitives such as all-reduce and all-to-all. Contrary to the impressions one might get from hobbyist single-device inference, the bigger constraint on inference at scale is not memory bandwidth but intra-node and network latency.

An even simpler approximation can be obtained by assuming that the number of layers \( L \) in a dense model scales as \( L \propto \nparam^{1/4} \), a scaling relationship that holds approximately for the dense models trained by \cite{hoffmann2022training}, for example. In this case, the scaling in number of parameters holding all other variables fixed simplifies to

\begin{equation}
    \label{eq:inverse-sqrt-latency-scaling}
    \text{token-to-token latency} \propto \sqrt{\nparam}
\end{equation}

\begin{figure}[h]
\centering
  \includegraphics[width=0.6\linewidth]{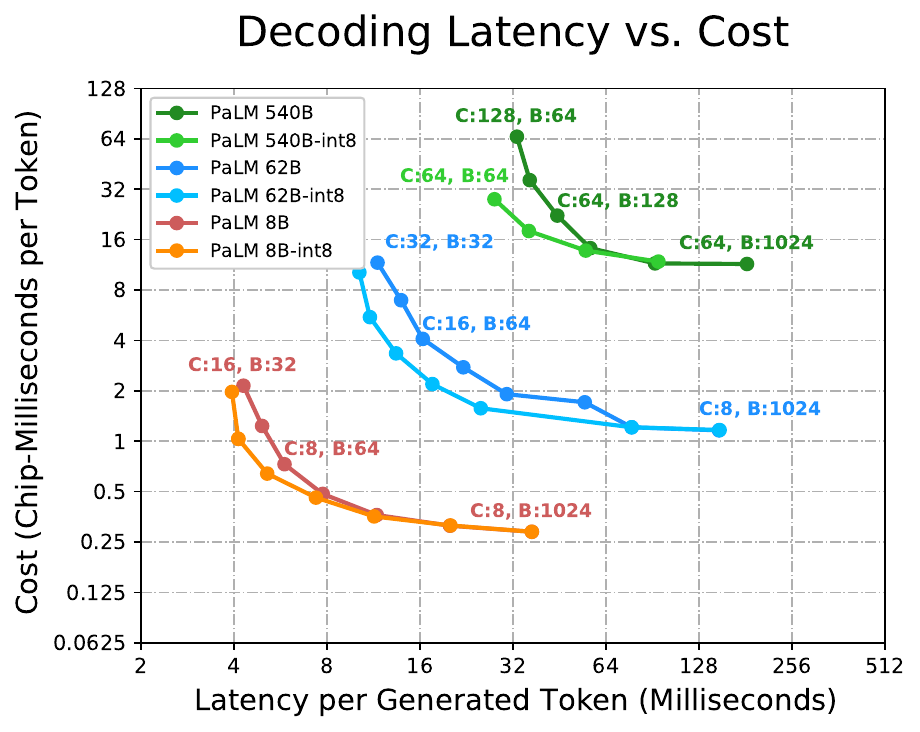}
  \caption{The cost per token versus token latency Pareto frontiers for PaLM 8B, 62B and 540B obtained by \cite{pope2022efficiently} on TPU v4 clusters. The best possible latencies for each model size approximately match the square root scaling predicted by our toy model.}
  \label{fig:esti-decode-cost-plot}
\end{figure}

This simple rule of thumb agrees with empirical data about dense models surprisingly well. For example, \cite{pope2022efficiently} obtain best latency figures of 4 ms, 13 ms and 32 ms per token for PaLM 8B, 62B and 540B respectively, as can be seen in Figure \ref{fig:esti-decode-cost-plot}, which agree with square root scaling of token latency in parameter count almost exactly.

We see a similar agreement when looking at more recent data on NVIDIA hardware. Taking Llama 3 8B at 400 tokens/second as a baseline, the square root model predicts 135 tokens/second for a 70B dense model and 56 tokens/second for a 405B dense model, which according to \cite{ArtificialAnalysis2024} are consistent with the speeds at which leading providers on NVIDIA hardware have been able to serve these models. Overall, the toy model performs remarkably well given how simple it is and how many important effects it neglects.

\section{Model description}
\label{sec:model_description}

We now build up the complete model we'll use in the rest of the paper on top of the more simplified model from Section \ref{sec:model-stage-2}. Here are the improvements we'll make:

\begin{enumerate}
    \item Take attention operations into account so that we can use the model to make useful predictions about long-context inference. This means factoring in both the arithmetic side (the quadratic complexity) and the memory bandwidth side (KV cache reads and storage).

    \item Relax the assumption that the model must be a dense model so that we can handle mixture-of-experts models.

    \item Use the more precise all-reduce and all-to-all latency approximations from \cite{NCCL} instead of the rough approximation we used earlier.

    \item Drop the assumption that network bandwidth is infinite and take network bandwidth constraints into account alongside network latency.

    \item Instead of modeling only tensor and data parallelism, also allow for pipeline and expert parallelism.
\end{enumerate}

We'll do these in order in the next five subsections before presenting the final equations of our model in a self-contained format.

\subsection{Handling the attention operations}

There are three points we must take into account when it comes to attention operations of a vanilla, decoder-only Transformer:

\begin{enumerate}
    \item The arithmetic required for taking the inner products of the query and key vectors.
    \item The memory bandwidth and memory capacity required for storing the KV cache.
    \item The fact that some extra GPU communication might be needed if the attention operations are substantial enough to require being distributed across multiple devices.
\end{enumerate}

The calculations become somewhat more complex for more recent innovations such as the \textit{multi-head latent attention} (MLA) mechanism from \cite{deepseekai2025deepseekv3technicalreport}. We will address this specific variant of attention after covering the vanilla case, though it's difficult to give a comprehensive review of all the ways in which the attention mechanism may be modified and all the ways these modifications would alter the calculations in this section.

To handle the arithmetic required for the attention operations, we add \( 4 \nlayer \nhead \dhead \ell_{\text{input}} b \flop \) of arithmetic operations that must be completed on top of the weight multiply-accumulates that give \( \approx 2 \nparam \, \flop \) of arithmetic per token. Here, \( \dhead, \nhead \) stand for the dimension of a single attention head and the number of total attention heads, \( \ell_{\text{input}} \) stands for input length, and \( b \) is the batch size.

The size of the KV cache is in general given by \( \nkvhead \dhead \nlayer \ell_{\text{input}} b \cdot p \) where \( \nkvhead \) is the number of distinct key \textit{and} value heads (so double-counting each head to count both key and value vectors). As before, \( p \) is the precision of the model parameters. We add this on top of the parameter reads from HBM that we're already taking into account. If we're willing to accept performance degradation, we can also compress the KV cache using various lossy methods, but for our model we will assume these methods are not used.

For the final point, the exact amount of extra communication needed will in general depend on the partitioning strategy used to divide the attention computation across multiple GPUs. For instance, \cite{pope2022efficiently} notes that partitioning the attention operation across the batch dimension incurs a cost of two additional all-to-all operations. However, the picture is complicated by the fact that when the KV cache is sufficiently small, one can instead partition the attention operations across the head dimension (i.e. assigning different heads to different GPUs) which does not incur the aforementioned extra communication cost. Because an accurate model of this would be rather complicated, in the remainder of this article we will ignore the extra communication cost arising from the attention operations.

Now, let's turn to addressing the case of multi-head latent attention. For our purposes, this variant of the attention mechanism is essentially equivalent to grouped-query attention except with increased arithmetic cost during decoding. Explicitly, the equivalence may be seen as follows:

\begin{enumerate}
    \item Multi-head latent attention uses a low-rank factorization of the key and value projections and caches the intermediate activations, also called "latent vectors", as a way of achieving KV cache compression. During decoding, one of the low-rank matrices appearing in this factorization are absorbed into the query projection (for the keys) and into the post-attention projection (for the values) respectively.

    The net effect is that the effective head dimensionality for computing the arithmetic cost of attention goes up from \( \dhead \) to \( \dlatent \), which is typically meaningfully larger. For instance, DeepSeek-V3 (\cite{deepseekai2025deepseekv3technicalreport}) uses \( \dhead = 128 \) and \( \dlatent = 512 \), raising the arithmetic cost of attention by a factor of \( 4 \) relative to a vanilla Transformer.

    \item In return for this increased arithmetic cost, the KV cache is compressed significantly: instead of having to cache key and value vectors each of size \( \nhead \dhead \) per input position, the vectors that need to be cached are just the latents, which are only of size \( \dlatent \) each. Since \( \dhead \ll \dlatent \ll \nhead \dhead \) in typical MLA implementations, this represents a huge reduction in the size of the KV cache even though it raises the arithmetic cost of attention.
\end{enumerate}

As we will later see, this memory IO versus arithmetic tradeoff makes MLA models unusual among Transformers in that their inference is prone to be arithmetic bound, rather than memory IO bound, in long contexts where attention dominates the cost of inference.

\subsection{Generalizing to mixture-of-experts models}

To generalize to mixture-of-experts models, we need to make appropriate changes to the arithmetic cost and memory reads necessary for a forward pass. First, to handle the arithmetic cost of a mixture-of-experts forward pass, we decompose the total number of parameters (excluding embedding parameters) of the model as follows:
\begin{equation}
    \nparam = N_{\text{attention params}} + N_{\text{feedforward params}} + N_{\text{unembedding params}}
\end{equation}
For a dense model, the arithmetic cost of a forward pass is simply \( 2 \nparam \flop \) per batch element. However, for a mixture-of-experts model with a sparsity factor of \( s = \nexpert/\nactiveexpert \) (where \( \nexpert \) is the total number of experts and \( \nactiveexpert \) is the number of experts that are active for any given token), the arithmetic cost is instead \( 2 \nactiveparam \flop \) per batch element, where
\begin{equation}
    \nactiveparam = N_{\text{attention params}} + \frac{N_{\text{feedforward params}}}{s} + N_{\text{unembedding params}}
\end{equation}
We only divide the number of feedforward parameters by \( s \) because the attention and unembedding parameters are shared across experts. This gives us the number of FLOP required for a forward pass.

We now move on to handle the memory bandwidth cost. Unlike for a dense model, we might not necessarily read all of the parameters of a mixture-of-experts model from memory in a given forward pass, because the batch might be small enough for some parameters to remain inactive across the entire batch. If we assume that expert activations are independently and uniformly distributed\footnote{Note that this assumption would be violated for the "shared experts" of MoE variants such as DeepSeek-V3. We still stick to the independent and uniform routing case because it's the most natural one, but our setup would need modifications before being applied to MoEs for which this assumption is violated.}, then the chance that any given expert will be inactive for any specific batch element is \( 1 - 1/s \), and the chance that it will be inactive for \textit{all} batch elements is \( (1 - 1/s)^b \) where \( b \) is the batch size as before.

Therefore, on average we expect the number of active experts to be \( \nexpert \cdot (1 - (1 - 1/s)^b) \), and therefore averaging across many batches, we expect the number of parameters read from memory to be
\begin{equation}
    N_{\text{params read}} = N_{\text{attention params}} + \left(1 - \left( 1 - \frac{1}{s} \right)^b \right) \cdot N_{\text{feedforward params}} + N_{\text{unembedding params}}
\end{equation}
When \( b \gg s \), we'll have \( (1 - 1/s)^b \approx 0 \), so in this case as expected we can approximate without much error that all of the model parameters will be read from memory during a single forward pass.

\subsection{Improving the latency approximation}
\label{sec:better-latency-approx}

To make our latency estimates more rigorous, we use the information from the tuning code in the official NCCL repository at \cite{NCCL}. Based on the information in the file ``src/graph/tuning.cc", we can deduce the following approximate latency expression for an all-reduce when using a tree topology:
\begin{equation}
    \label{eq:allreduce-latency-formal}
    6.8 \mus + (1.2 \mus) \cdot \left( \frac{n_{\text{ranks}}}{n_{\text{nodes}}} - 1 \right) + (10 \mus) \cdot \log_2(n_{\text{nodes}})
\end{equation}
Here, \( n_{\text{ranks}} \) is the number of ranks (or GPUs) participating in the all-reduce, while \( n_{\text{nodes}} \) is the number of participating nodes. Of course, while this expression is quite precise, our modeling so far has only used information about the overall number of GPUs participating in an all-reduce, and we've not taken the distribution of those GPUs across different nodes into account. To do this, we'll simply use another approximation: if our inference instance has a total of \( \ngpu \) GPUs distributed across \( N_{\text{nodes}} \) nodes, we'll assume that each all-reduce involves \( \sqrt{\ngpu} \) GPUs across \( \sqrt{N_{\text{nodes}}} \) nodes. While this is still an approximation, it's significantly more accurate than the latency expression we used in Section \ref{sec:model-stage-2}.

\subsection{Taking network communication time into account}

Having taken network latency into account, we now move on to consider network bandwidth constraints. This won't affect the order of magnitude of our decoding speed estimates for flagship NVIDIA GPUs, because in practice network bandwidth time only dominates latency for batch sizes larger than the critical batch size on the H100. Specifically, for realistic per-GPU batch sizes of up to 100 and a model dimension of \( 10,000 \), network bandwidth limits can add

\begin{equation}
    \frac{(2 \text{ bytes}) \cdot 10,000 \cdot 100 \cdot (8-1)}{8 \cdot 112 \text{ GB/s}} \approx 16 \mus
\end{equation}

of extra time to each all-reduce operation on a DGX H100 if LL, the lowest latency communication algorithm of NCCL with \( 112 \text{ GB/s} \) of all-reduce algorithm bandwidth, is used. This is not small enough to be negligible, so we have to take it into account in a more realistic model.

To determine how much time is taken up by network communication, we need to answer the following questions:

\begin{enumerate}
    \item How many bytes do we need to all-reduce?
    \item How many nodes and ranks are participating in each all-reduce?
    \item To what extent can the network communication be overlapped by memory reads or arithmetic?
\end{enumerate}

For the first question, assuming that we have an all-reduce after each matrix multiplication, we just need to add up the number of entries of each output matrix in a single layer and multiply this by the precision of the model's parameters and the number of layers. This is relatively straightforward: the QKV matrix is of size \( (1 + 2/g) \nhead \dhead b \), the post-attention projection result is of size \( \dmodel b \), the first and second matrix multiplications in the feedforward block then contribute \( \dff b \) and \( \dmodel b \) respectively. Here, \( g \) is the attention group size and \( b \) is the batch size. One caveat to keep in mind is that depending on whether the architecture uses an activation scheme such as SwiGLU, there might be more than one output matrix having the dimensions \( \dff \times b \).

Therefore, for a model with a typical Transformer architecture (i.e. no SwiGLU), the total number of bytes we need to all-reduce in a single forward pass using tensor paralellism is 
\begin{equation}
    \text{bytes reduced} = \left( (1 + 2/g) \nhead \dhead + 2 \dmodel + \dff \right) \cdot b \cdot \nlayer \cdot p
\end{equation}
where \( b \) is the batch size, \( \nlayer \) is the number of layers, and \( p \) is the precision of the model parameters.

For (2), we've already answered this question in Section \ref{sec:better-latency-approx}: the number of participating ranks is \( \sqrt{\ngpu} \) and the number of participating nodes is \( \sqrt{N_{\text{nodes}}} \). Therefore, we can express the total amount of intra-node and inter-node communication that needs to happen as follows:
\begin{align}
    \text{inter-node network reads} &= 2 \cdot (\sqrt{N_{\text{nodes}}} - 1) \cdot \text{bytes reduced} \\
    \text{intra-node network reads} &= 2 \cdot (\sqrt{N_{\text{GPU}}/N_{\text{nodes}}} - 1) \cdot N_{\text{nodes}} \cdot \text{bytes reduced}
\end{align}
The expressions here arise from the fact that reducing \( X \) bytes across \( R \) participants requires \( 2X(R-1) \) bytes to be read, in general. Treating each node as a single participant gives the first expression, and then treating each node independently and assuming that an all-reduce is performed across all the GPUs in each node gives the second expression.

Now, to obtain the time taken for the communication, we only need to divide these by the intra-node and inter-node communication bandwidth available, so we obtain the final expression
\begin{equation}
    \text{network communication time} = \frac{\text{inter-node network reads}}{\ngpu \cdot \text{inter-node bandwidth}} + \frac{\text{intra-node network reads}}{\ngpu \cdot \text{intra-node bandwidth}}
\end{equation}
For example, for the H100 the intra-node bandwidth would be 450 GB/s. This is half of the reported NVLink bandwidth of 900 GB/s because the reported bandwidth is not full-duplex: it takes both reads and writes into account, while our network reads expression is counting only network reads. Separately, the inter-node bandwidth for e.g. a cluster consisting of DGX H100 systems would be 50 GB/s, because each DGX H100 system comes together with 8 ConnectX-7 VPI cards, each of which supports 50 GB/s of full-duplex bandwidth. Since the number of GPUs per DGX H100 system is also equal to 8, we have one card per GPU, resulting in a network bandwidth of 50 GB/s per GPU.

For the final question about overlapping, we will conservatively assume that network communication cannot be overlapped with memory reads and arithmetic. This is generally going to be true when using off-the-shelf primitives from NCCL, and overlapping the network communication in an all-reduce with computation using simple methods (e.g. splitting the all-reduce into smaller pieces so that computation can begin on the pieces that are processed first while later pieces are still being reduced) seems to require taking the latency hit from the all-reduce multiple times. This is acceptable during training, but not during inference, because in inference the forward passes are required to be much faster than during training. This inability to split the all-reduce into smaller pieces without multiplying the latency per all-reduce is the reason behind our conservative assumption here.

A significant implication of taking network communication into account is that the critical batch size is no longer the optimal batch size choice in general. This is because unlike parameter reads that can be amortized over a bigger batch, network communication per batch scales linearly with the batch size. Fast inference often requires reducing this communication time by scaling down the batch size beyond the critical batch size as we will see later.

\subsection{Allowing pipeline and expert parallelism}

The model from Section \ref{sec:model-stage-2} only took tensor and data parallelism into account. In this section, we discuss how we can incorporate pipeline and expert parallelism in a simple way into the basic model.

We address pipeline parallelism (PP) first. Ignoring PP is not a major issue in most cases, because if we have a pipeline with \( \npp \) stages, we'll generally need \( \npp \) micro-batches in flight in order to avoid pipeline bubbles. This means pipeline parallelism slices the batch into smaller pieces just like data parallelism (DP) does. When training a model there's a solid reason to do this: DP requires gradient all-reduces after each forward and backward pass, and using both PP and DP at the same time can economize on this bandwidth cost, essentially because the communication cost of both PP and DP scales linearly with their parallelism degree, while the overall cluster size (and total communication bandwidth available) scales with their product.

However, when we're doing inference, there is no backward pass and subsequent gradient update, which means DP requires no communication whatsoever past the initial routing of tokens to the right DP copies. This eliminates the central reason to use PP over DP in training, and means the only advantage of PP left in inference is that it enables sharding model weights across multiple devices and thereby help the model weights fit into GPU memory at the expense of introducing some communication overhead at pipeline boundaries.

While this can be useful in theory, on NVIDIA hardware efficient inference setups are generally not constrained by availability of memory. An easy way to see this for H100s is that reading the contents of 80 GB of HBM alone would take \( \approx 24 \text{ ms} \) at the theoretical peak HBM bandwidth of \( 3.3 \text{ TB/s} \), so the Pareto fronts of optimal cost versus speed must be identical for H100s with infinite memory and with 80 GB of memory past a speed of \( (24 \text{ ms/token})^{-1} \approx 41 \text{ tokens/sec} \), and in practice even sooner once we take into account network bandwidth limits and more realistic sustained HBM bandwidth values.\footnote{This calculation assumes we're not using speculative sampling so that each parameter has to be read from HBM once per generated token.} As a result, there's generally no reason to use pipeline parallelism when data parallelism is available as an alternative.

We can model pipeline parallelism by lowering the effective batch size in our model to \( \microbatch = b/\npp \), adding \( \npp - 1 \) sequential peer-to-peer communications of size \( \dmodel \cdot \microbatch \) each for the processing of each token that can be paralellized across \( \ngpu/\npp \) GPUs, and relaxing the memory capacity constraint to allow for weight and KV cache sharding across PP copies. We would then loop over a grid of values for possible PP degrees and see which one gives the lowest token latency for a fixed instance and batch size.

Having handled PP, let's move on to expert parallelism (EP). This method of parallelism can only be used on the feedforward blocks because experts generally share the same attention block, so using EP requires attention and feedforward blocks to be parallelized differently, a detail that's important to handle correctly in an inference model. In general, we'll be using tensor parallelism on the post-attention projection, so we'll end up with a residual stream tensor that's scattered across the \( \dmodel \) dimension. For expert parallelism, we need to perform the following two communication steps:

\begin{enumerate}
    \item Each expert parallel GPU needs to receive the batch elements that are routed to it: this is the \textit{all-to-all dispatch}. In the worst case this will require each vector of size \( \dmodel \) to be transmitted \( r = \min(\nep, \nactiveexpert) \) times, but in practice it could be substantially less if expert activations are correlated and we place experts likely to fire together on the same EP copies. So we get an upper bound on the communication needed: \( r \cdot \dmodel \cdot \microbatch \), which we can approximately think of as an all-to-all operation across \( r \) ranks of a tensor of size \( \dmodel \cdot \microbatch \).

    \item After each expert has finished its computation, we need to once again scatter the results across the model dimension to proceed: this is the \textit{all-to-all receive}. This operation is simply the reverse flow of the previous operation logically, so we can also model it as an all-to-all with exactly the same type. 
\end{enumerate}

This simple estimate shows why expert parallelism is enormously cheaper than tensor parallelism:

\begin{enumerate}
    \item When \( \nep \ll \nactiveexpert \), even if we make the most pessimistic assumption about how active experts distribute over EP copies, the communication volume of expert parallelism is at most identical to that of 1D tensor parallelism: two all-to-alls versus a single all-reduce of tensors of identical size over identically many ranks. If expert activations are suitably correlated, we can often make this communication volume substantially lower by optimizing expert placement across EP ranks.

    \item When \( \nep \gg \nactiveexpert \), \textit{the communication volume of expert parallelism doesn't scale with the number of expert parallel ranks, only with the number of active experts}. So parallelizing up to the upper limit of \( \nep = \nexpert \) costs little to no additional communication\footnote{It might cost additional latency depending on the network topology, however.}, though we incur some cost from having to move some of this communication to InfiniBand instead of NVLink for large EP degrees. Still, this is enormously more favorable than what we would have to deal with with large TP degrees in a 2D TP setup.
\end{enumerate}

So an assumption that's generically safe is that whenever the instance size \( \ngpu \) is below the number of experts \( \nexpert \) in an MoE model, we would try to maximally use expert parallelism. If \( \ngpu > \nexpert \), then we can split the remaining \( \ngpu/\nexpert \) degrees of parallelism over TP and PP in whatever way is optimal for latency, holding \( \ngpu \) and batch size constant. This is what we do in our inference economics model.

\subsection{Equations for the final model}
\label{sec:equations-final-model}

To conclude this section, we present the complete equations of our final model in the 2D tensor parallel case and under the assumption of no pipeline or expert parallelism. This presentation assumes a vanilla decoder-only Transformer architecture that uses ReLU (or similar) activations: small changes need to be made to the memory and network reads/writes if a SwiGLU activation (\cite{shazeer2020glu}) is used to take the additional weight matrix in the feedforward blocks into account. We also assume that the KV cache and the model weights fit into the HBM of all GPUs combined across the cluster: if this is not true, the model returns an infinite token latency.

The code for the paper contains a more general model which relaxes the 2D TP and no PP or EP assumptions and does casework to handle them according to the discussion from Section \ref{sec:model_description}, but we do not include the equations of the model in this more general case in the paper.

\begin{align}
    \text{token latency} &= \text{network and kernel time} + \max \left( \frac{\text{bytes read}}{\ngpu \cdot \text{memory bwd}}, \frac{\text{total FLOP}}{\ngpu \cdot \text{GPU FLOP/s}} \right) \\
    \text{network and kernel time} &= \nlayer \nreduce t_{\text{kernel latency}} + \text{network comm time} \\
    \text{bytes read} &= \text{weight precision} \cdot \text{parameters read} + \text{activation precision} \cdot \text{activations read} \\
    \text{parameters read} &= N_{\text{attention params}} + \left(1 - \left( 1 - \frac{1}{s} \right)^b \right) \cdot N_{\text{feedforward params}} + N_{\text{unembedding params}} \\
    \text{activations read} &= \text{KV cache elements read} + \text{matmul input and outputs read} \\
    \text{KV cache elements read} &= \nkvhead \dhead \nlayer \ell_{\text{input}} b \\
    \text{matmul input and outputs read} &= \nlayer \cdot b \cdot (2\dmodel + (1 + 2/g) \nhead \dhead + \nhead \dhead + 2\dmodel + 2\dff) \\
    \text{total FLOP} &= b \cdot (\text{pointwise FLOP} + \text{attention FLOP}) \\
    \text{attention FLOP} &= (4 \flop) \cdot \nlayer \nhead \dhead \ell_{\text{input}} \\
    \text{pointwise FLOP} &= (2 \flop) \cdot \left( N_{\text{attention params}} + \frac{N_{\text{feedforward params}}}{s} + N_{\text{unembedding params}} \right) \\
    \text{network comm time} &= \nlayer \nreduce \treduce + \frac{\text{inter-node reads}}{\ngpu \cdot \text{inter-node bandwidth}} + \frac{\text{intra-node reads}}{\ngpu \cdot \text{intra-node bandwidth}} \\
    \treduce &= 6.8 \mus + (1.2 \mus) \cdot \left( \sqrt{\frac{N_{\text{GPU}}}{N_{\text{nodes}}}} - 1 \right) + (10 \mus) \cdot \log_2(\sqrt{N_{\text{nodes}}}) \\
    \text{inter-node reads} &= 2 \cdot (\sqrt{N_{\text{nodes}}} - 1) \cdot \text{bytes reduced} \\
    \text{intra-node reads} &= 2 \cdot (\sqrt{N_{\text{GPU}}/N_{\text{nodes}}} - 1) \cdot \sqrt{N_{\text{nodes}}} \cdot \text{bytes reduced} \\
    \text{bytes reduced} &= \left( (1 + 2/g) \nhead \dhead + 2 \dmodel + \dff \right) \cdot b \cdot \nlayer \cdot \text{activation precision}\footnotemark \\
    \nreduce &= 4
\end{align}
\footnotetext{This assumes a feedforward block with a vanilla activation function. For more complex activations such as SwiGLU, it might be necessary to add another \( \dff \) term to the expression inside the parantheses.}

\section{Inference economics of example models}
\label{sec:inference_well_known}

Having presented the model, we now move on to discussing the inference economics of some well-known example models running on H100s at a price of two dollars per hour per H100 SXM. The H100 SXM has the following specs which will be relevant for our calculation:

\begin{itemize}
    \item The tensor cores can perform \( 10^{15} \text{ FP16/s}, \, 2 \cdot 10^{15} \text{ FP8/s}, \, 2 \cdot 10^{15} \text{ INT8/s} \) under ideal conditions. We assume in practice sustained arithmetic throughput caps out at \( 70 \% \) of these claimed numbers due to thermal throttling, unmodeled nonlinearities in the Transformer, et cetera.
    \item The HBM has a size of \( 80 \text{ GB} \) and a bandwidth of \( 3.3 \text{ TB/s} \). We assume that in practice sustained bandwidth caps out at \( 75 \% \) of this theoretical maximum.
    \item Per-GPU bidirectional NVLink bandwidth of \( 450 \text{ GB/s} \): \( 450 \text{ GB/s} \) of reads and \( 450 \text{ GB/s} \) of writes adds up to the \( 900 \text{ GB/s} \) in official specs. If using the LL (low-latency) protocol for NCCL primitives, which is necessary to get the good latency numbers discussed in Section \ref{sec:model_description}, these bandwidths are further divided by a factor of \( 2 \). We also assume a node size of \( 8 \) GPUs, though some recent developments might allow this to be increased for Ampere and Hopper GPUs.
    \item The kernel launch latency (e.g. for CUTLASS kernels doing GEMM operations) is around \( 4 \mus \).
    \item The base latency for launching GPU collectives in NCCL is around \( 6.8 \mus \) when using the LL protocol, which we assume cannot be eliminated and has to be paid for each GPU collective that's launched sequentially.
\end{itemize}

\begin{figure}[h]
    \centering
    \begin{minipage}{0.5\textwidth}
        \centering
        \includegraphics[width=\textwidth]{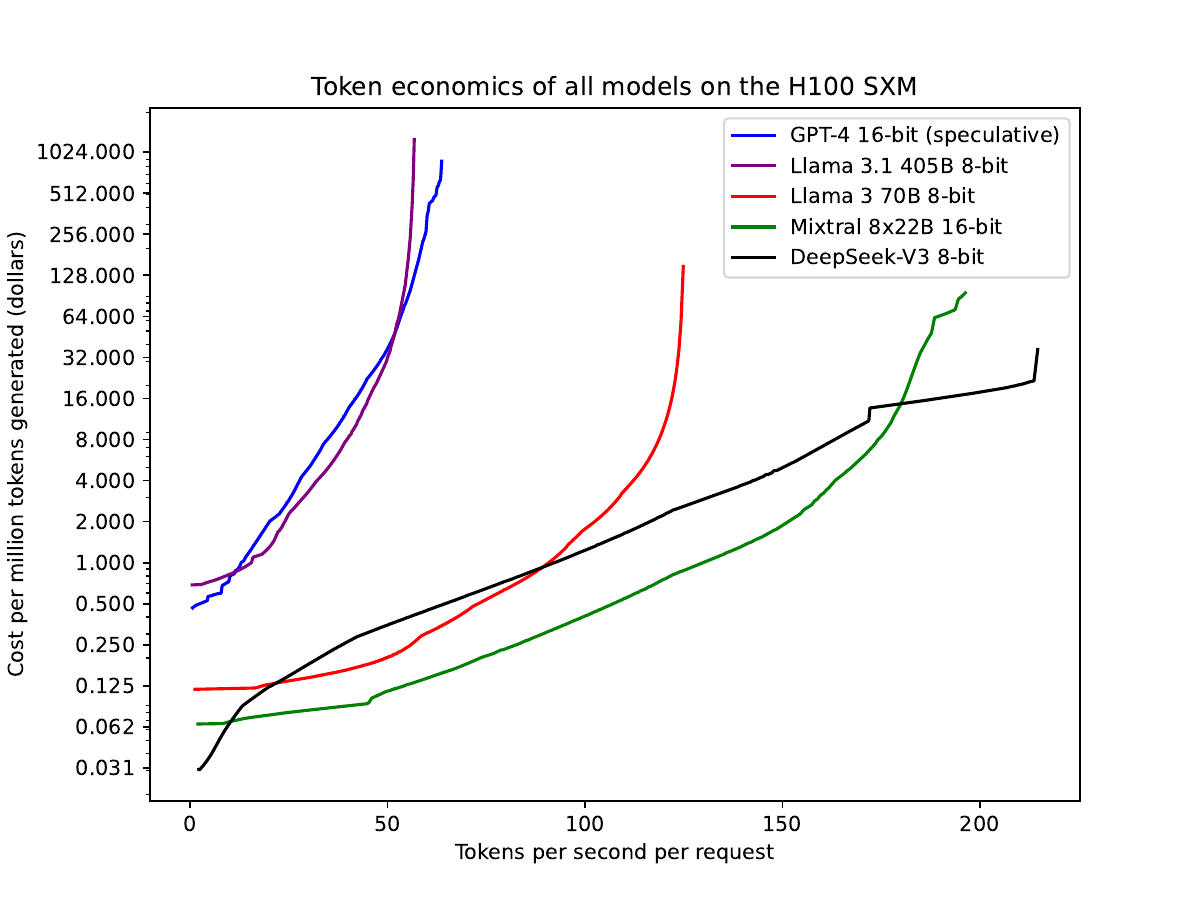}
        \caption{Without speculative decoding.}
        \label{fig:main-token-economics-plot}
    \end{minipage}\hfill
    \begin{minipage}{0.5\textwidth}
        \centering
        \includegraphics[width=\textwidth]{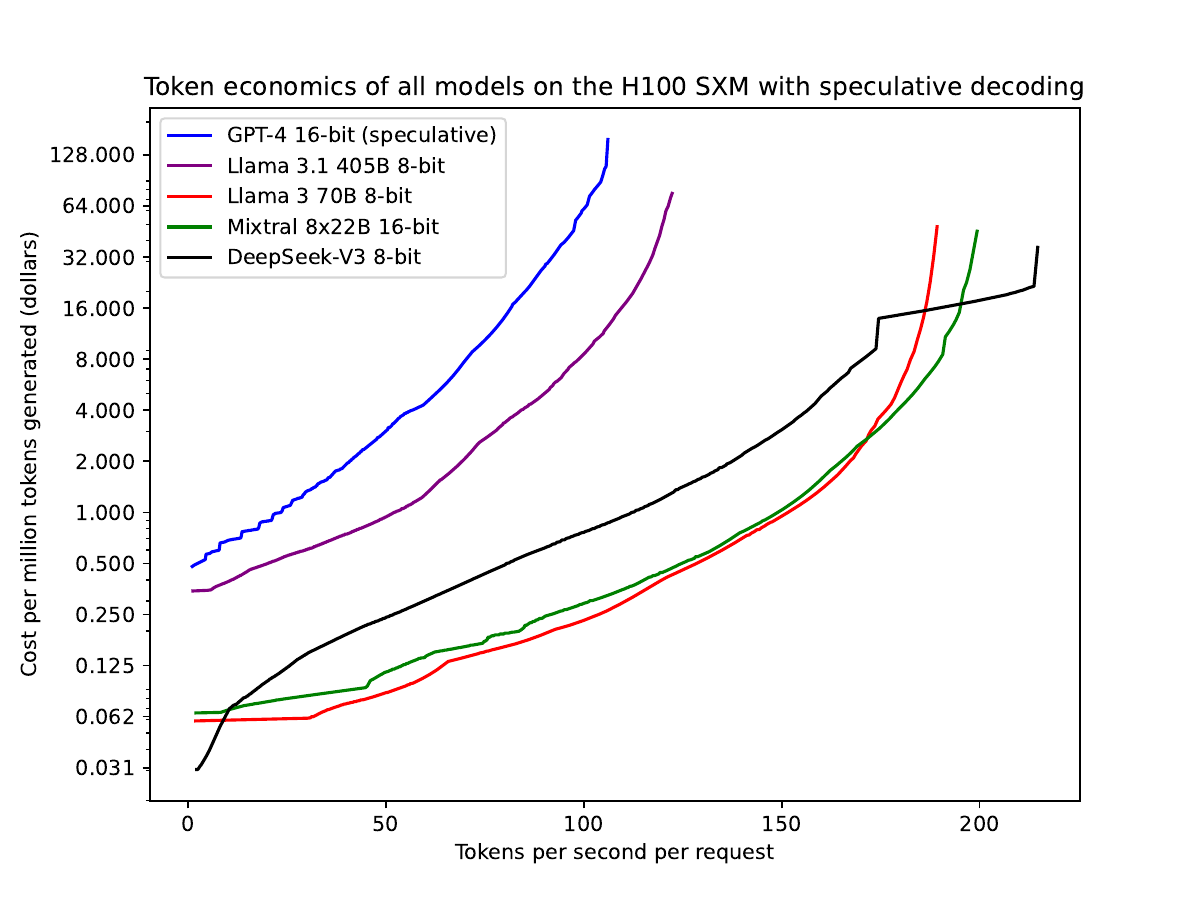}
        \caption{With speculative decoding.}
        \label{fig:main-token-economics-plot-with-spec-dec}
    \end{minipage}
    \caption{The token economics for GPT-4, Llama 3.1 405B, Llama 3 70B\protect\footnotemark, DeepSeek-V3, and Mixtral 8x22B predicted by our model. In the right plot, we assume speculative decoding by Llama 3 8B (\cite{leviathan2023fastinferencetransformersspeculative}), while the left plot corresponds to the token economics of serving the base models. As before, we assume an effective rental price of 2 USD/hr for each H100 SXM.}
\end{figure}

Using these numbers for the model in Section \ref{sec:model_description}, we obtain the results in Figures \ref{fig:main-token-economics-plot} and \ref{fig:main-token-economics-plot-with-spec-dec}.

\footnotetext{We assume that Llama 3 70B weights are quantized to an 8-bit format such as FP8 or INT8, while the activations remain at a 16-bit precision.}

As before, we also provide the maximum achievable tokens per second per request for each model, along with the optimal inference instance size at which this optimal performance is achieved. These results can be found in Table \ref{tab:max-tokens-per-second-by-model-final}.

\begin{table}[h]
\centering
\begin{tabular}{lcc}
\toprule
\textbf{Model name} & \textbf{Maximum tokens per second} & \textbf{Optimal instance size (GPUs)} \\ \midrule
Mixtral 8x22B (16-bit)       & 199                         & 125 \\ 
Llama 3 70B (8-bit)          & 189                         & 24 \\ 
Llama 3.1 405B (8-bit)       & 122                         & 48 \\
GPT-4 (16-bit, speculative)  & 106                         & 460 \\
DeepSeek-V3 (8-bit)          & 215                         & 14 \\
\bottomrule
\end{tabular}
\caption{Maximum tokens per second and the optimal instance size required to achieve the maximum according to our final model from Section \ref{sec:model_description} using H100 SXM GPUs. We assume speculative decoding is used, with the speculative decoder being taken as Llama 3 8B with a token acceptance probability of \( 80 \% \). Note that these numbers disregard cost and only try to maximize tokens per second per request, so e.g. they assume a batch size of \( 1 \) even for mixture-of-experts models to reduce the total amount of memory reads.}
\label{tab:max-tokens-per-second-by-model-final}
\end{table}

An important observation to be made in Figure \ref{fig:main-token-economics-plot} is that even assuming 8-bit quantization for Llama 3 70B is not enough for the observed price-performance point from the leading provider to be to the left of the model's Pareto frontier. This suggests that the observed performance is not achievable when doing naive autoregressive inference using Llama 3 70B alone; it's necessary to use speculative decoding or other similar techniques that reduce the number of calls to the model.\footnote{It's also possible that API providers have more optimized GPU collective kernels than we've assumed thus far, but for the moment we're assuming this is not the case.} We'll later see that the latency gains from speculatively decoding Llama 3 70B using Llama 3 8B as the approximator model are significant: an optimal setup can effectively cut the latency per token by a factor of \( 2 \). This is sufficient for our model to be consistent with the data.

Now that we have a method for computing Pareto frontiers, we can ask the question of which point on the Pareto frontier of the API provider makes sense to choose given the preferences of the provider's clients. To perform this analysis in a simplified way, suppose that tokens generated at different speeds are perfect substitutes for each other, and the marginal value of a token to a representative customer scales with \( (\text{tokens per second})^{\alpha} \) for some \( \alpha \geq 0 \): tokens which are generated faster are worth more, all else equal. In this case, the customer will choose to purchase at the speed that maximizes

\begin{equation}
\label{eq:customer-preference}
    \frac{(\text{tokens per second})^{\alpha}}{\text{price per million tokens}}
\end{equation}

Therefore, in equilibrium, we will observe that all API providers serve their models at the point on the cost curves from Figure \ref{fig:main-token-economics-plot} that maximizes Equation \ref{eq:customer-preference}. Different values of \( \alpha \) give rise to different outcomes, and empirically we observed that we need to pick \( \alpha \approx 3 \) to make the data from \cite{ArtificialAnalysis2024} consistent with our model. Using this value of \( \alpha \), we present the implied efficient inference setups in Table \ref{tab:tokens-per-second-by-model-efficient}.

\begin{table}[h]
\centering
\begin{tabular}{lcccc}
\toprule
\textbf{Model name} & \textbf{Tokens per second} & \textbf{Cost per million tokens} & \textbf{Instance size (GPUs)} & \textbf{Batch size} \\ \midrule
Mixtral 8x22B (16-bit)       & 128 & \$0.54  & 23  & 197 \\ 
Llama 3 70B (8-bit)          & 107 & \$0.27  & 7   & 136 \\ 
Llama 3.1 405B (8-bit)       & 61  & \$1.31  & 8   & 58 \\
GPT-4 (16-bit, speculative)  & 61  & \$4.53  & 100 & 213 \\
DeepSeek-V3 (8-bit)          & 116 & \$1.10  & 54  & 247 \\
\bottomrule
\end{tabular}
\caption{The efficient inference setups according to our final model from Section \ref{sec:model_description} that maximize Equation \ref{eq:customer-preference}, assuming \( \alpha = 4 \) and H100 SXM GPUs. We assume speculative decoding is used, with the speculative decoder being taken as Llama 3 8B with a token acceptance probability of \( 80 \% \).}
\label{tab:tokens-per-second-by-model-efficient}
\end{table}

Instead of fixing a GPU and looking at what happens when we serve different models on it, we can also compare different GPUs with each other to see how much cost savings we get from using newer GPUs such as the H100 compared to older ones such as the V100. For this comparison, we need to assume some rental prices, and we do this by taking the lowest rates we can find on \href{https://nebius.ai}{nebius.ai} (which correspond to orders renting many GPUs at once for a long period) and assuming that a third of the price corresponds to markups charged by the cloud provider (this factor doesn't matter for the comparison since it's the same for all GPUs). This gives the following rental prices:

\begin{itemize}
    \item \textbf{H100 SXM:} 2.1 USD/hr
    \item \textbf{A100 SXM:} 1.5 USD/hr
    \item \textbf{V100 SXM:} 0.42 USD/hr
\end{itemize}

\begin{figure}[h]
\centering
  \includegraphics[width=0.8\linewidth]{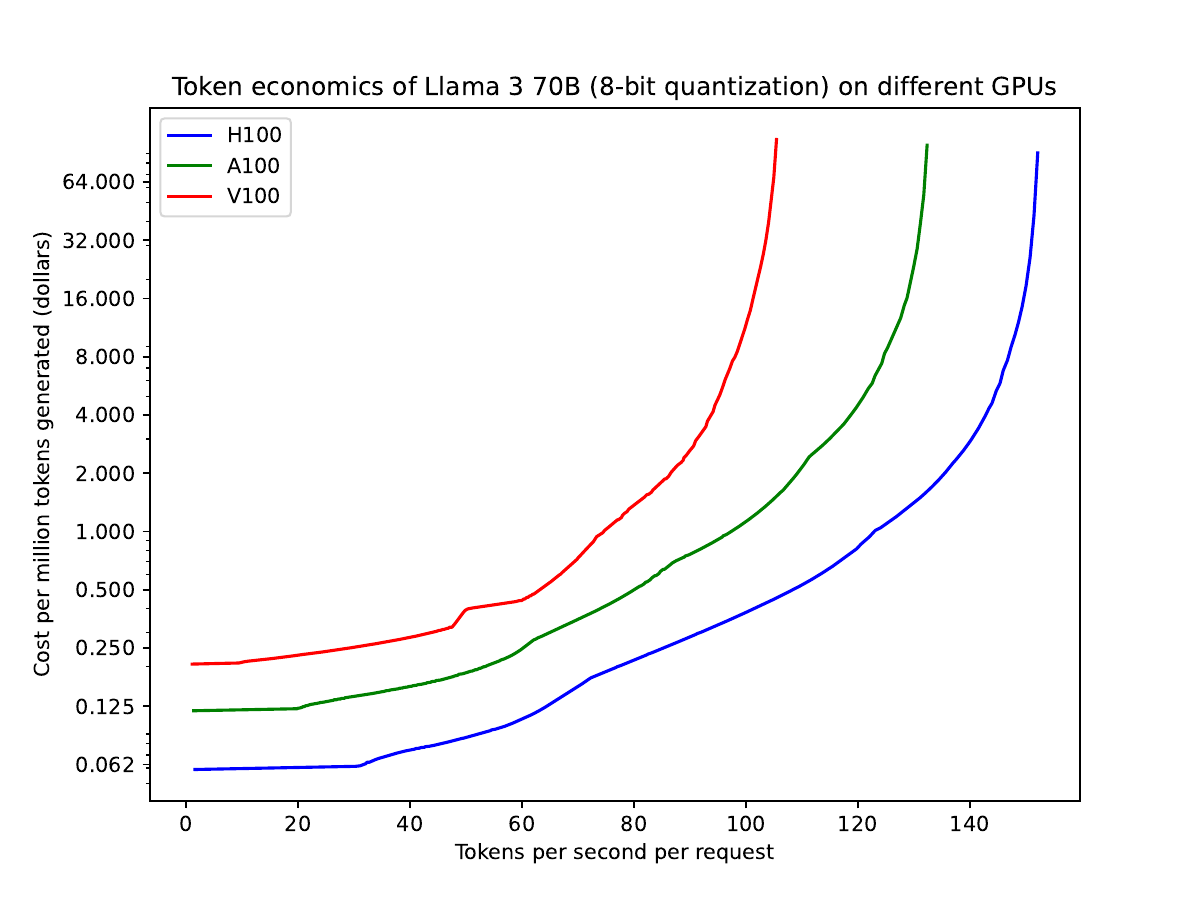}
  \caption{The token economics for Llama 3 70B with weights quantized to 8-bit precision on the V100, the A100 and the H100 (all of them using the SXM form factor). At large batch sizes, the Ampere and Hopper architectures benefit from their tensor cores supporting faster 8-bit precision arithmetic.}
  \label{fig:gpus-token-economics-plot}
\end{figure}

Choosing inference of Llama 3 70B with 8-bit quantized weights as our benchmark to judge the performance of these GPUs, we get the results presented in Table \ref{tab:max-tokens-per-second-by-gpu-final} and Figure \ref{fig:gpus-token-economics-plot}.

\begin{table}[h]
\centering
\begin{tabular}{lcc}
\toprule
\textbf{GPU name} & \textbf{Maximum tokens per second for Llama 3 70B inference} & \textbf{Optimal instance size (GPUs)} \\ \midrule
H100 SXM       & 152                         & 24 \\ 
A100 SXM       & 132                         & 32 \\ 
V100 SXM       & 105                         & 102 \\
\bottomrule
\end{tabular}
\caption{Maximum tokens per second and the optimal instance size required to achieve the maximum according to our final model from Section \ref{sec:model_description} on Llama 3 70B with the weights quantized to 8-bit precision. Different rows correspond to different GPUs.}
\label{tab:max-tokens-per-second-by-gpu-final}
\end{table}

Figure \ref{fig:gpus-token-economics-plot} and Table \ref{tab:max-tokens-per-second-by-gpu-final} suggest the following rough approximation across a wide range of inference speeds: with each new generation of GPUs from NVIDIA, the tokens per second achieved at a fixed cost per token goes up by around \( 25 \% \), and the instance size required for inference on the Pareto frontier at a fixed cost per token falls by a factor of \( 2 \). It remains to be seen whether new generations of GPUs such as the B100 will continue this trend.

\subsection{The impact of speculative decoding}

Our inference economics model so far assumes that each new token generated to respond to a request requires a forward pass of the model that we're serving for inference. This is a reasonable assumption due to the sequential nature of text generation: because responses are generated one token at a time, and because later tokens causally depend on earlier tokens, it might naively seem as if we can't do better than this. However, there are techniques that allow us to circumvent this restriction, and the most popular out of them is speculative decoding (\cite{leviathan2023fastinferencetransformersspeculative}). 

The idea of speculative decoding is to use a smaller model which can be inferenced more quickly as a "speculator" or "approximator" to the larger model and use a rejection sampling technique to ensure that the tokens thus sampled from the smaller model have the same probability distribution as we would obtain if we had directly sampled from the larger model. More specifically, if \( P \) is the larger model and \( Q \) is the smaller model used as a speculator (both of which are using the same vocabulary and tokenizer), the procedure for sampling tokens looks like the following:

\begin{enumerate}
    \item Fix an integer \( \gamma \geq 1 \). 
    \item Sample \( \gamma \) tokens \( x_1, x_2, \ldots, x_{\gamma} \) from \( Q \). Store the probabilities \( q_i \) of the \( i \)th token under \( Q \) conditional on previous context.
    \item Do a single forward pass on all of \( x_1, x_2, \ldots, x_{\gamma} \) with \( P \) to determine their probabilities \( p_1, p_2, \ldots, p_{\gamma} \) of being sampled from \( P \) conditional on previous context.
    \item For each \( i = 1, 2, \ldots, \gamma \), accept the \( i \)th token with probability \( \min(1, p_i/q_i) \) and reject the token otherwise. 
    \item Let \( j \) be the index of the first rejected token. Discard the tokens \( x_j, x_{j+1}, \ldots, x_{\gamma} \) and resample \( x_j \) from the probability distribution whose mass function is proportional to \( \max(0, P-Q) \). 
    \item Append \( x_1, x_2, \ldots, x_j \) to the previous context and go back to step (2).
\end{enumerate}

It's easy to prove that this procedure is equivalent to sampling the tokens from \( P \) directly. The advantage lies in the fact that if the acceptance rate is high enough for \( j \) to be suitably large in step (5), we can take a large value of \( \gamma \) and ideally only call \( P \) approximately once per \( \gamma \) tokens generated. If forward passes of \( Q \) are much faster than those of \( P \), the latency advantage we would get from this is apparent. We do incur an additional arithmetic cost from this, as disregarding failed speculative branches each token still has to be seen once by \( P \) and once by \( Q \), but when \( Q \) is much cheaper than \( P \) this cost is negligible.

If we model token acceptance events as independent and identically distributed, so that each token is accepted with a fixed probability \( \alpha \), the number of tokens generated in each step follows a truncated geometric distribution whose expected value is given by \( V = (1-\alpha^{\gamma})/(1-\alpha) \). If the latency per forward pass of the models is assumed to be constant and given by \( t_P, t_Q \), the average latency per token under speculative decoding will be given by

\begin{equation}
    \label{eq:spec-dec-latency}
    \text{mean latency per token} = \frac{\text{time per iteration}}{V} = \frac{(1 - \alpha) (t_P + \gamma t_Q)}{1 - \alpha^{\gamma}}
\end{equation}

\( \gamma \) is an integer-valued parameter we have the freedom to choose, and we can make the choice that minimizes latency for a fixed value of \( \alpha \). We can estimate \( t_P, t_Q \) given the architectures of the two models using our inference economics model from earlier. Therefore, the only information specific to speculative decoding we need to know is \( \alpha \), the acceptance probability per token. If we have access to both \( P \) and \( Q \), we can estimate this quantity for a given prompt \( y \) by the following procedure:

\begin{enumerate}
    \item Sample a sequence of tokens \( x_1, x_2, \ldots, x_n \) from \( P \) as a response to (or continuation of) the prompt \( y \).
    \item For a randomly chosen index \( i \), sample one token \( z_i \)  from \( Q \) conditional on the context \( y; x_1, x_2, \ldots, x_{i-1} \).
    \item Compute \( p = P(z_i \mid y; x_1, x_2, \ldots, x_{i-1}) \) and \( q = Q(z_i \mid y; x_1, x_2, \ldots, x_{i-1}) \). Then, we expect \( \alpha = \mathbb E[\min(1, p/q)] \).
    \item Using (3), we can obtain more precise estimates of \( \alpha \) by repeating the sampling and computation steps in (2) and (3) for many prompts and many positions in the responses to these prompts, then averaging all of the values \( \min(1, p/q) \).
\end{enumerate}

Full access to the models \( P \) and \( Q \) is not required for this procedure: it's enough to have access to an API which provides the log probabilities of generated tokens. Many API providers offer this information in their responses to requests, e.g. \href{https://together.ai}{together.ai} and \href{https://octo.ai}{octo.ai}. We can therefore estimate the acceptance rate that would be achieved if a big model served by an inference provider was speculatively decoded using a smaller model that they also serve via API. Typical acceptance rates we've obtained when doing this in the Llama model series at sampling temperature equal to 1 range from \( 0.7 \) to \( 0.9 \), depending on the exact prompts and the pair of models chosen, and these findings agree with the range of acceptance rates reported for different model families where the two models are around an order of magnitude apart in size in \cite{leviathan2023fastinferencetransformersspeculative}.

\begin{figure}[h]
\centering
  \includegraphics[width=0.8\linewidth]{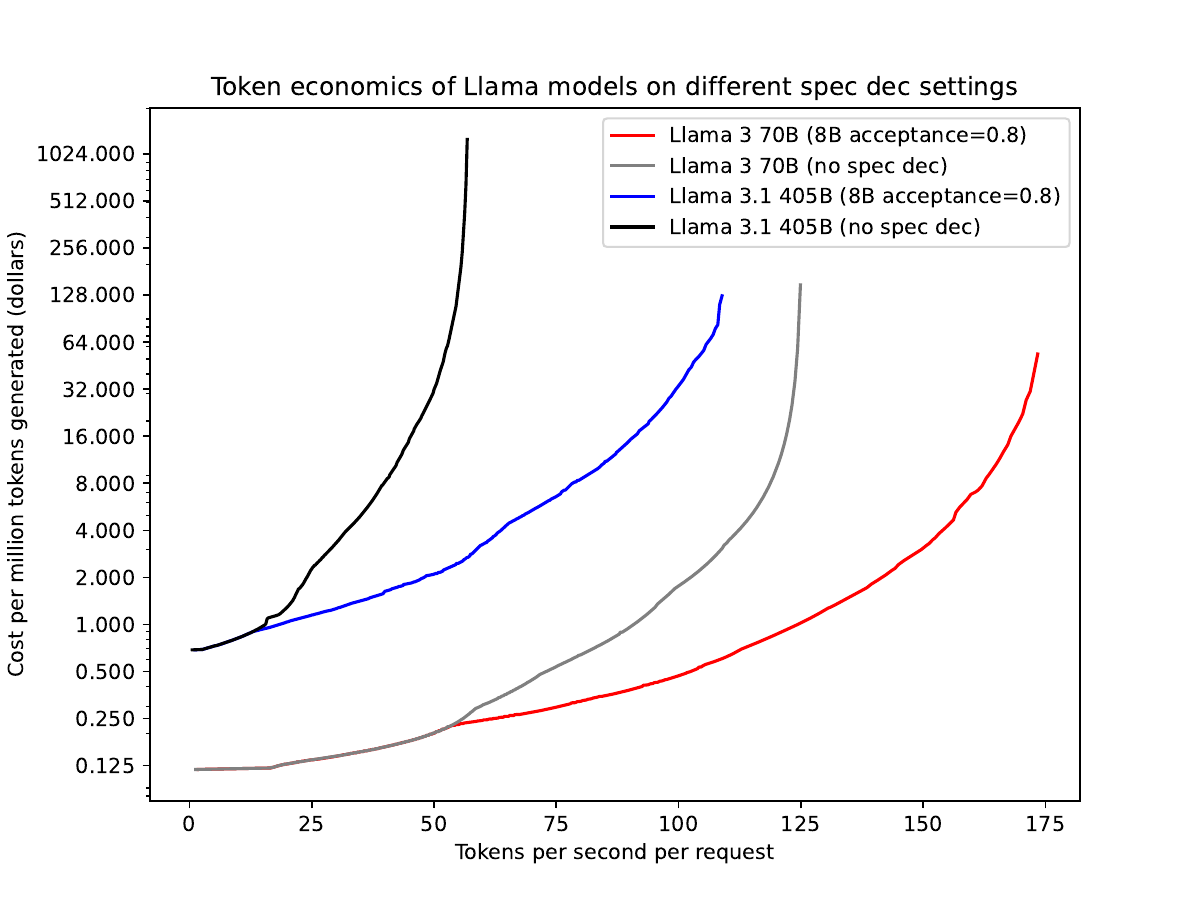}
  \caption{The token economics for Llama 3 70B and Llama 3.1 405B when both models are speculatively decoded with a model having Llama 3 8B's architecture as the speculator. We provide the curves without speculative decoding for purposes of comparison. Here, all three models involved are served at a 16-bit precision, so no quantization is being employed.}
  \label{fig:spec-dec-token-economics-plot}
\end{figure}

Using \( \alpha = 0.8 \) as our reference value for purposes of illustration, we get the results in Figure \ref{fig:spec-dec-token-economics-plot} and Table \ref{tab:tokens-per-second-by-spec-dec-efficient}. Overall, the gains from speculative decoding at an \( 80 \% \) token acceptance rate are substantial: a \( 66 \% \) increase in throughput for Llama 3 70B and a doubling of throughput for Llama 3.1 405B at a fixed cost per token. The size of the effect means that it's difficult for API providers not using speculative decoding to compete with ones that do on latency or pricing.

\begin{table}[h]
\centering
\begin{tabular}{lcccc}
\toprule
\textbf{Weight format} & \textbf{Tokens per second} & \textbf{Cost per million tokens} & \textbf{Instance size (GPUs)} & \textbf{Batch size} \\ \midrule
Llama 3 70B            & 69  & \$0.52  & 8   & 127 \\
Llama 3 70B (S)        & 95  & \$0.51  & 6   & 73  \\ 
Llama 3.1 405B         & 35  & \$6.54  & 31  & 80  \\ 
Llama 3.1 405B (S)     & 71  & \$7.33  & 23  & 26  \\
\bottomrule
\end{tabular}
\caption{The efficient inference setups for Llama 3 70B and Llama 3.1 405B with or without speculative decoding with Llama 3 8B as the speculator at a \( 80 \% \) token acceptance rate. (S) following the name of a model means we assume speculative decoding is used for the model in question. As before, we choose the efficient setups by maximizing Equation \ref{eq:customer-preference}, assuming \( \alpha = 3 \) and H100 SXM GPUs. Note that these models are not quantized and remain at 16-bit weight precision.}
\label{tab:tokens-per-second-by-spec-dec-efficient}
\end{table}

\subsection{The impact of quantization}

Quantization is one of the primary techniques used by API providers to reduce inference costs and token-to-token latency. In general, weight quantization helps because it reduces memory bandwidth usage and accelerates arithmetic by shifting to lower precision formats supported by tensor cores, while activation quantization helps by reducing the size of the KV cache and by reducing network bandwidth usage.

\begin{figure}[h]
\centering
  \includegraphics[width=0.8\linewidth]{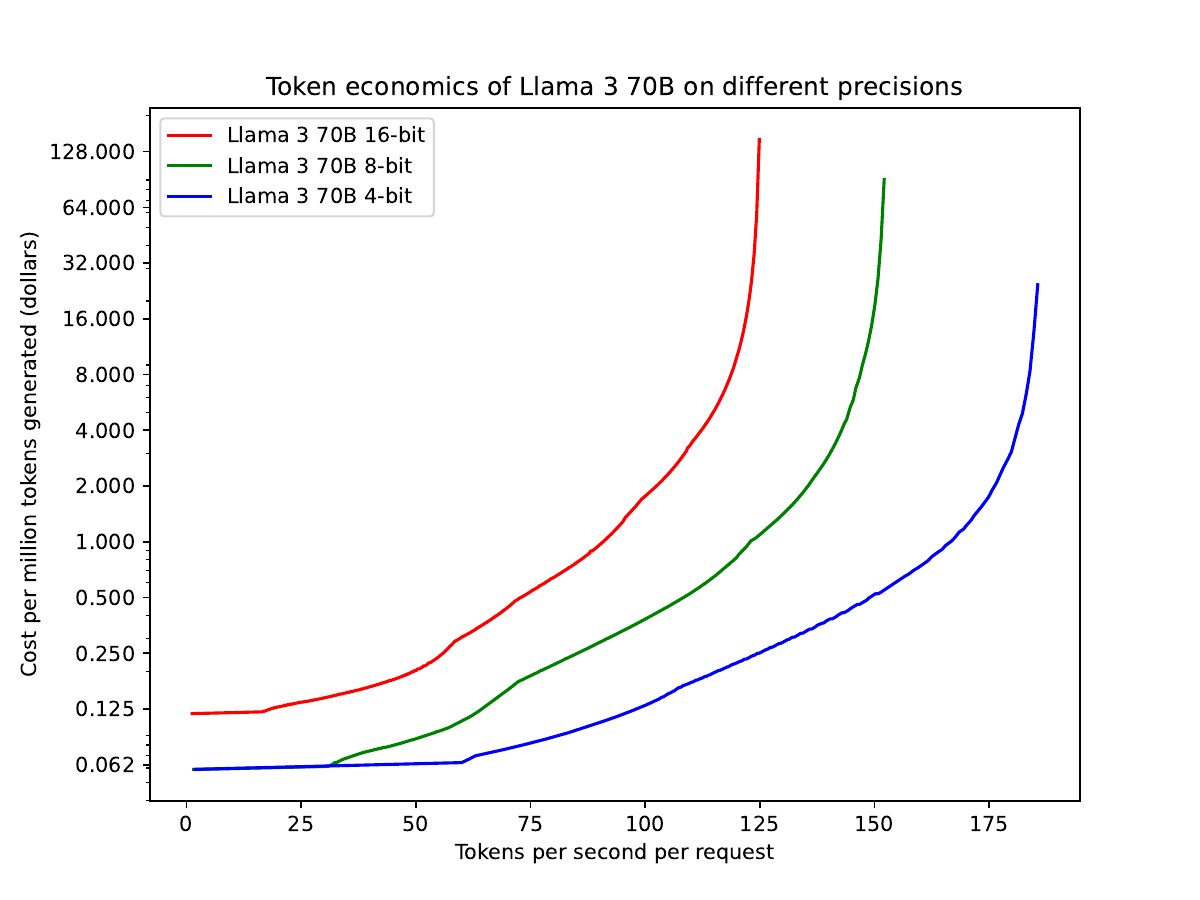}
  \caption{The token economics for Llama 3 70B with weights quantized to different precisions, holding activation precision fixed at 16 bits.}
  \label{fig:quant-token-economics-plot}
\end{figure}

Once again taking Llama 3 70B as our baseline, we provide some results about how much quantization helps with inference. Our simplified model from Equation \ref{eq:minimum-token-latency-approx} would predict that each halving of weight precision should give a \( 2^{1/3} - 1 \approx 26 \% \) increase in maximum tokens per second. The results from the more detailed model can be found in Figure \ref{fig:quant-token-economics-plot} and Table \ref{tab:tokens-per-second-by-quant-efficient}, and they show that this simple rule of thumb actually works quite well.

\begin{table}[h]
\centering
\begin{tabular}{lcccc}
\toprule
\textbf{Weight format} & \textbf{Tokens per second} & \textbf{Cost per million tokens} & \textbf{Instance size (GPUs)} & \textbf{Batch size} \\ \midrule
4-bit weights       & 122 & \$0.23  & 4   & 90  \\ 
8-bit weights       & 99  & \$0.37  & 7   & 109 \\ 
16-bit weights      & 83  & \$0.70  & 13  & 136 \\
\bottomrule
\end{tabular}
\caption{The efficient inference setups for Llama 3 70B with weights quantized to different precisions according to our final model from Section \ref{sec:model_description} that maximize Equation \ref{eq:customer-preference}, assuming \( \alpha = 4 \) and H100 SXM GPUs.}
\label{tab:tokens-per-second-by-quant-efficient}
\end{table}

\subsection{The impact of long context lengths}

When the context length is not long enough for KV cache reads or attention arithmetic to make a substantial contribution to the total cost of inference, the main difference between input and output tokens is that input tokens can be processed in parallel and as a result generally have much more forgiving latency constraints. Indeed, in many cases just a ping to a target server might take on the order of 100 milliseconds, so trying to shave off milliseconds from the processing that happens on the server side is usually not worth it. This leads to a much better average arithmetic utilization when processing input tokens compared to output tokens.

For instance, at \( 100 \% \) utilization on an H100, the cost per million tokens of Llama 3 70B at 16-bit precision would be around \( \$0.11 \), implying an utilization rate of \( \approx 17 \% \) for this precision in Table \ref{tab:tokens-per-second-by-quant-efficient}. During prefill (the processing of input tokens before any output tokens are generated) we probably expect this to be \( 50 \% \) or more, suggesting that input tokens should be priced 3 times cheaper than output tokens. This ratio matches empirical data about e.g. OpenAI's pricing of their input tokens versus output tokens for GPT-4o (\cite{openai_pricing}).

In contrast, when the context length is so long that attention dominates the cost of inference, the lower utilization during decoding occurs due to the need to read the KV cache every time a new token is generated. The size of the KV cache per batch element can be expressed as \( 2 \nlayer \nhead \dhead \linput/g \) activations, while the amount of arithmetic per token can be expressed as \( 4 \nlayer \nhead \dhead \linput \flop \). Therefore, the arithmetic intensity in this regime is \textit{at most}

\[ \frac{4 \nlayer \nhead \dhead \linput \flop}{\text{precision} \cdot 2 \nlayer \nhead \dhead \linput/g} = \frac{2g \flop}{\text{precision}} \]

if speculative decoding or other similar methods to decode multiple tokens for one read of the KV cache are not employed. This means that in sufficiently long context lengths, the differential pricing of input and output tokens under an ideal pricing scheme would depend almost entirely on the attention group size \( g \) of the architecture, with all other details being essentially irrelevant. When grouped-query attention is not used, so that \( g = 1 \), the best utilization that can be achieved on NVIDIA hardware will be below \( 1 \% \). For this reason, GQA is critical for efficient long-context inference regardless of other architectural details.

In practice, because most API providers use fixed prices for input and output tokens, their pricing strategy can only be an approximation to the ideal strategy suitable for the demand pattern that they face for their service, with details such as the ratio of input to output tokens in a typical long-context request being of particular importance.

\begin{figure}[h]
\centering
  \includegraphics[width=0.8\linewidth]{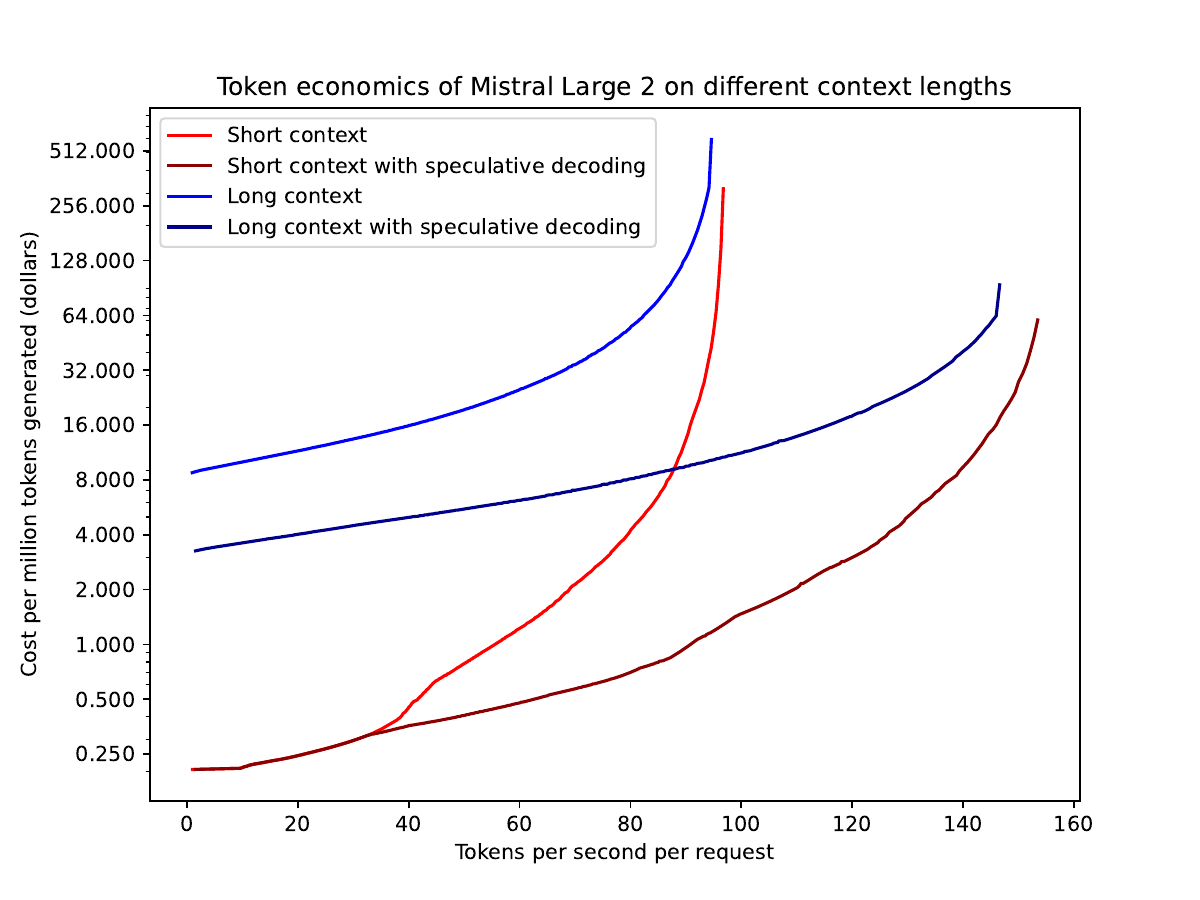}
  \caption{The token economics for Mistral Large 2 at short context lengths and long context lengths (100K). We assume the speculative decoder is a hypothetical model with a variant of Llama 3 8B's architecture using multi-query attention with an \( 80 \% \) acceptance rate for proposed tokens.}
  \label{fig:mistral-token-economics-plot}
\end{figure}

As an interesting case study, we compare the short and long context token economics of Mistral Large 2, a dense transformer with 123 billion parameters and an attention group size of 12.  The predictions of the inference model for these settings can be found in Figure \ref{fig:mistral-token-economics-plot}.

Mistral Large 2 has a per-token KV cache size of around \( 360 \text{ KB} \). At a context length of 100K, this means each generated token requires \( 10^5 \cdot 360 \text{ KB} = 36 \text{ GB} \) of KV cache reads without speculative decoding. Since the model size itself is only \( 244 \text{ GB} \), KV cache reads become a bigger bottleneck than parameter reads at batch sizes larger than \( 7 \). In addition, at \( 3.3 \text{ TB/s} \) of memory bandwidth per H100 GPU with GPU time costs at 2 USD per hour, the price of the memory read time required for KV cache reads per million tokens can be calculated as

\[ \frac{36 \text{ PB/million output tokens}  \cdot \$2/\text{hour}}{3.3 \text{ TB/s}} = \$6.06 \text{ per million output tokens} \]

Since this price is a lower bound on the actual price per million tokens in most situations\footnote{It's possible to get around this to some extent using clever scheduling methods, for example by batching long context requests with short context ones and keeping the tensor cores occupied during the memory reads for the KV cache. However, these methods only yield limited gains, and our model assumes a fixed context length over the entire batch.}, the difference between the short and long context Pareto frontiers ends up being quite large.

\begin{figure}[h]
\centering
  \includegraphics[width=0.8\linewidth]{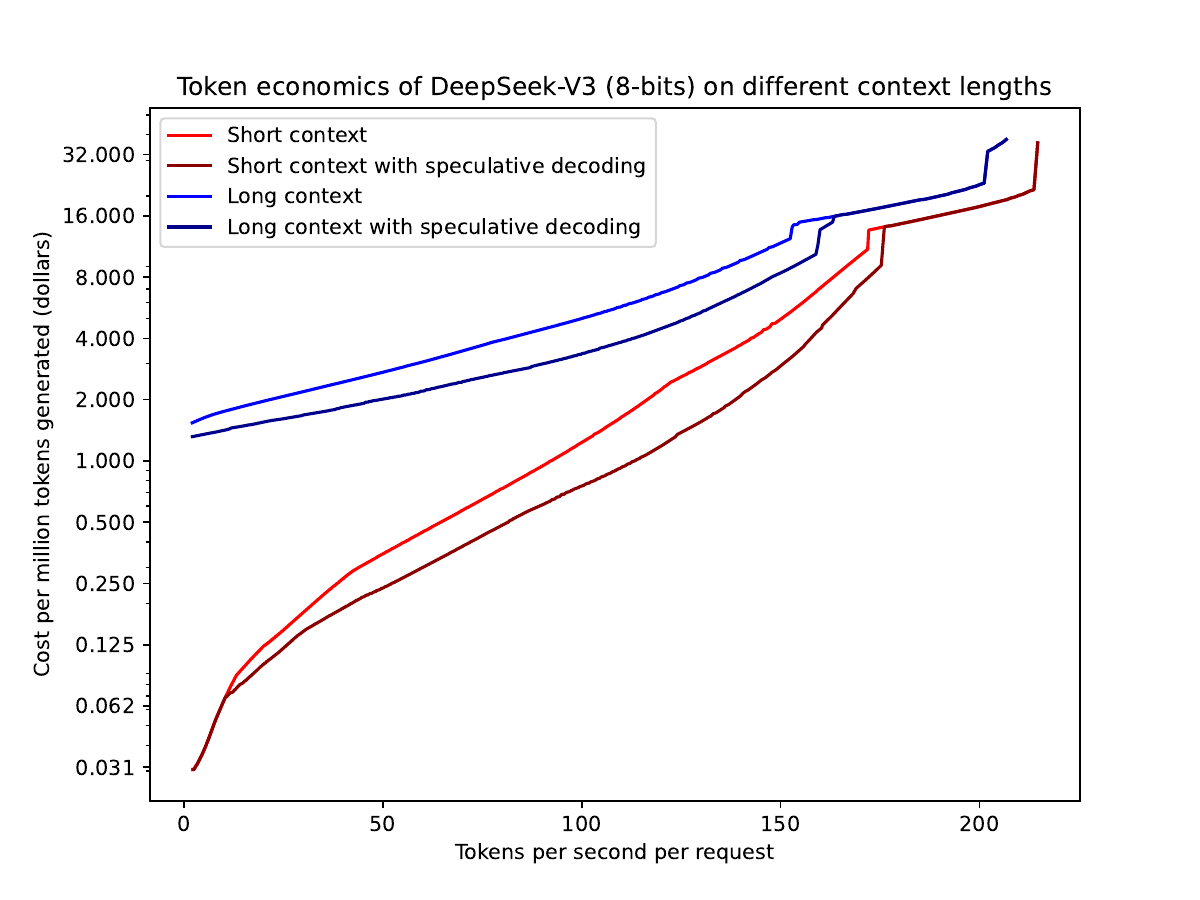}
  \caption{The token economics for DeepSeek-V3 (8-bit weight quantization) at short context lengths and long context lengths (100K). We assume the speculative decoder is a hypothetical model with a variant of Llama 3 8B's architecture using multi-query attention with an \( 80 \% \) acceptance rate for proposed tokens.}
  \label{fig:deepseek-v3-token-economics-plot}
\end{figure}

It's instructive to contrast this with the long-context behavior of a highly sparse MLA model such as DeepSeek-V3. With a per-token KV cache size of \( \approx 60 \text{ KB} \), we expect DeepSeek-V3 to be around 6 times cheaper than Mistral Large 2 if we're doing inference slowly. This turns out to roughly check out in our model as seen in Figure \ref{fig:deepseek-v3-token-economics-plot}.

\section{Conclusion and important takeaways}

We list some of the most important takeaways from our analysis below. 

\subsection{Takeaways for short-context inference}

\begin{enumerate}
    \item The most important bottleneck to fast decoding in short contexts is \textbf{latency}, with network and NVLink latency being the biggest culprits on NVIDIA hardware. Even small all-reduce operations can cost tens of microseconds of pure latency.

    If latency constraints didn't exist, then we could in principle increase the serial speed of token generation by four times indefinitely for each doubling of cost per token (\cite{steinhardt2022}), because each matrix multiplication can be parallelized along three different dimensions, so increasing our number of GPUs by 8 times with 4 times more 2D tensor parallelism and 2 times more pipeline or data parallelism would only increase both memory and network bandwidth usage by a factor of 2. Any modeling of real-world token economics has to begin by understanding how latency causes this basic logic to break down.

    \item A rough rule of thumb is that the speed at which a dense model with \( N \) parameters can be served for inference scales as \( \propto 1/\sqrt{N} \): there's an inverse square root law. This can be empirically verified using data from sources such as \cite{ArtificialAnalysis2024} but it also drops out of the theoretical analysis that we perform on a toy model in Section \ref{sec:theoretical-analysis} if we make some reasonable assumptions about model aspect ratio scaling.\footnote{This agreement shouldn't be overstated, as in practice there are multiple factors that would lead us to expect latency to increase faster with model size that Section \ref{sec:theoretical-analysis} does not consider. The fact that the scaling law nevertheless empirically holds shows that the factors that this model does not consider appear to roughly cancel out in practice up to an overall constant factor in the scaling law.}

    \item Higher memory bandwidth does speed up inference, but the increase is slow in short contexts: a doubling of memory bandwidth alone can only be expected to deliver a reduction of token-to-token latency of around \( 20 \% \) or less. This is implied by toy models of inference such as in Section \ref{sec:theoretical-analysis} and is also confirmed by empirical data: Groq LPUs have up to \( 80 \text{ TB/s} \) of memory bandwidth per chip compared to \( 3.3 \text{ TB/s} \) per H100 SXM, and as our model would predict, the latency gap between the fastest NVIDIA provider and Groq on \cite{ArtificialAnalysis2024} is generally a factor of 2 to 3.\footnote{Groq is able to overcome their very small amount of memory per chip (around 230 MB) by making extensive use of pipeline parallelism during inference. This allows them to approach the performance that would be attained by the same memory bandwidths at much higher amounts of on-chip memory, which is what the model in Section \ref{sec:theoretical-analysis} estimates.}

    \item Speculative decoding at plausible token acceptance rates has a large impact on token-to-token latency, possibly reducing it by a factor of $ 2 $ at a fixed cost per token in realistic situations. This is in spite of the added arithmetic cost of speculative decoding and the potential for speculated tokens to be rejected.

    \item Quantizing models is most useful when quantization allows the model to fit inside a discontinuous hardware boundary such as a single GPU or a single node. In general, reducing the precision of a model by a factor of \( 2 \) seems to reduce token latency by around \( 20 \% \), but the decrease in latency can be significantly more than this in the event that the inference setup is running up against a natural hardware boundary.

    \item Assuming that an inference provider is able to get good utilization during the prefill phase, the hardware utilization during short-context decoding can be anywhere from 3 to 5 times less than the one achieved during prefill at practical speeds.
 \end{enumerate}

 \subsection{Takeaways for long-context inference}

 \begin{enumerate}
     \item In leading open source models today which use a traditional attention mechanism (Llama series of models, Mistral Large 2, \textit{et cetera}), the use of limited grouped-query attention with group sizes ranging between \( 4 \) and \( 12 \) means that long-context inference is \textbf{always memory-bandwidth bound} due to KV cache reads. This does not depend on the batch size used for inference, because the size of the KV cache scales linearly with the batch size. Overcoming this bottleneck requires big group sizes (on the order of \( g = 100 \)), KV cache compression methods, or revisions to the attention mechanism itself.
     
     One such revision is \textit{multi-head latent attention} (MLA), used for example in DeepSeek-V3 (\cite{deepseekai2025deepseekv3technicalreport}). This variant of attention makes a significant tradeoff, raising arithmetic cost by many times in order to match the efficient KV cache size reduction of GQA with large group sizes without the same performance degradation. The large arithmetic cost of the attention mechanism of MLA models means their inference often becomes attention arithmetic bound at context lengths as short as ten thousand tokens.

     \item As a consequence of (1), increases in memory bandwidth are typically more important for long-context inference compared to short-context inference, as in longer contexts even increasing the batch size and slowing down inference cannot improve utilization when KV cache reads become the bottleneck.
 \end{enumerate}

Overall, the theoretical model we've developed accurately predicts many empirical facts about LLM inference economics without having been fit to the data or adjusted with "fudge factors" by hand. The framework has the potential to serve as a practical "roofline model" for understanding the fundamental limits of LLM inference performance.

\printbibliography

\end{document}